\documentclass{article} % For LaTeX2e
\usepackage{iclr2026_conference,times}

\usepackage{amsmath,amsfonts,bm}

\def\eqref#1{equation~\ref{#1}}
\def\1{\bm{1}}

\DeclareMathAlphabet{\mathsfit}{\encodingdefault}{\sfdefault}{m}{sl}
\SetMathAlphabet{\mathsfit}{bold}{\encodingdefault}{\sfdefault}{bx}{n}

\usepackage{hyperref}
\usepackage{url}
\usepackage{subcaption}
\usepackage{amsmath}
\usepackage{booktabs}
\usepackage{pifont}
\usepackage{amsmath}   % for general math environments
\usepackage{amssymb}   % includes \mathbb and other symbols
\usepackage{multirow}
\usepackage{enumitem}
\usepackage{xspace}  % For smart spacing after commands
\usepackage{colortbl}      % optional: if using row colors
\usepackage{arydshln}      % for \cdashline
\usepackage{graphicx}      % for \resizebox
\usepackage{amssymb}       % for symbols like \textsuperscript
\usepackage{caption}
\usepackage{subcaption}
\usepackage{wrapfig}
\usepackage[dvipsnames]{xcolor}
\usepackage{algorithm}
\usepackage{algpseudocode}   % for \Require, \Ensure, \For, \If
\usepackage{tabularx}
\usepackage{listings}

\title{Procedural Mistake Detection via \\ Action Effect Modeling}

\author{Wenliang Guo, Yujiang Pu,  Yu Kong \\
Michigan State University, USA\\
\texttt{\{guowenli,puyujian,yukong\}@msu.edu} \\
}

\newcommand{\eg}{e.g.\@\xspace}
\newcommand{\ie}{i.e.\@\xspace}

\iclrfinalcopy % Uncomment for camera-ready version, but NOT for submission.
\begin{document}

\maketitle

\begin{abstract}
Mistake detection in procedural tasks is essential for building intelligent systems that support learning and task execution. Existing approaches primarily analyze how an action is performed, while overlooking what it produces, i.e., the \textbf{action effect}. Yet many errors manifest not in the execution itself but in the resulting outcome, such as an unintended object state or incorrect spatial arrangement. To address this gap, we propose Action Effect Modeling (AEM), a unified framework that jointly captures action execution and its outcomes through a probabilistic formulation. AEM first identifies the outcome of an action by selecting the most informative effect frame based on semantic relevance and visual quality. It then extracts complementary cues from visual grounding and symbolic scene graphs, aligning them in a shared latent space to form robust effect-aware representations. To detect mistakes, we further design a prompt-based detector that incorporates task-specific prompts and aligns each action segment with its intended execution semantics. Our approach achieves state-of-the-art performance on the EgoPER and CaptainCook4D benchmarks under the challenging one-class classification (OCC) setting. These results demonstrate that modeling both execution and outcome yields more reliable mistake detection, and highlight the potential of effect-aware representations to benefit a broader range of downstream applications. \footnote{Project page: \url{https://wenliangguo.github.io/Mistake_Detection}}
\end{abstract}
\section{Introduction}
\label{sec:intro}

Mistakes are inevitable in procedural tasks, from cooking and assembly to medical procedures. While humans refine their skills through learning and experience, they often overlook subtle execution details that lead to unintended outcomes. For instance, a person might follow what seems to be the correct cutting motion, yet still produce irregular cucumber slices due to a slight misalignment in technique. These errors may not be immediately apparent from the action itself but become evident in the final result. To better support users in performing tasks, intelligent systems must therefore assess not only how actions are performed, but also what those actions ultimately produce.

Existing mistake detection approaches focus primarily on modeling the execution process, assessing correctness by analyzing motion patterns and action sequences. Previous works have explored such methods in structured tasks, such as assembly \citep{sener2022assembly101} and cooking \citep{wang2023holoassist, peddi2023captaincook4d}, often relying on frame-level classification or multimodal features. More recently, \citet{lee2024error} introduced prototype learning to evaluate execution similarity across instances, while \citet{huang2025modeling} utilized task graphs to predict all possible next actions and compared them with actual executions in the latent space. However, a shared limitation of these methods is the assumption that mistakes can be identified solely from the execution process, without verifying whether the final outcome aligns with the intended result.

\begin{figure}[t]
  \centering
  \begin{subfigure}{1.0\linewidth}
    \includegraphics[width=1.0\linewidth]{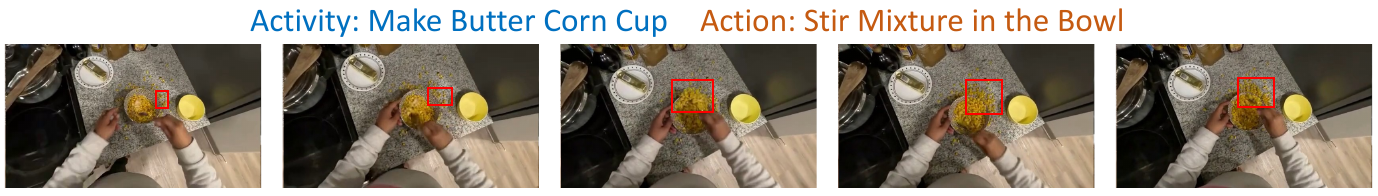}
    \caption{Positional Mistake, \ie, spill out mixture onto table.}
    \label{subfig:intro_1}
  \end{subfigure}
  \hfill
  \begin{subfigure}{1.0\linewidth}
    \includegraphics[width=1.0\linewidth]{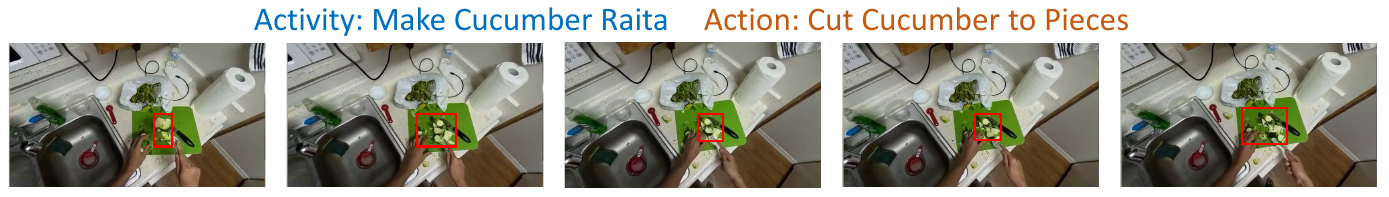}
    \caption{State mistake, \ie, cucumber is cut into unexpected shape.}
    \vspace{-4pt}
    \label{subfig:intro_2}
  \end{subfigure}
  \caption{Examples of two types of mistakes: object positional and state mistakes, in cooking scenario from CaptainCook4D dataset \citep{peddi2023captaincook4d}.}
  \vspace{-10pt}
  \label{fig:intro}
\end{figure}

In real-world scenarios, the execution may appear correct, yet minor deviations can lead to flawed outcomes. As illustrated in Figure \ref{subfig:intro_1}, an imprecise stirring position might resemble a correct motion, while the resulting spillage reveals the error. In Figure \ref{subfig:intro_2}, subtle differences in slicing technique may produce irregularly shaped slices. These examples underscore a critical limitation: mistake detection should account for not only the execution process but also the resulting action effect.

We address this challenge by casting mistake detection as a marginalization problem over latent variables including potential action effects. Concretely, the probability of a mistake is modeled as a joint function of both the execution of an action and its outcome. This probabilistic view naturally decomposes the task into three subproblems: identifying the action effect, interpreting the effect, and determining whether it reveals a mistake. Based on this formulation, we propose \textbf{Action Effect Modeling (AEM)}, a unified framework that enriches action representations with effect-aware features. AEM first selects an effect frame that best reflects the result based on both semantic relevance and visual quality. The action effect is then modeled from two complementary perspectives: \emph{object states}, which capture visual attributes such as shape and size changes, and \emph{spatial relationships}, which describe how objects are arranged relative to one another. 

To learn robust and interpretable effect-aware representations, AEM leverages multimodal supervision from vision-language models \citep{hurst2024gpt} through dual branches. The visual branch encodes grounded object features and positions, while a textual branch constructs symbolic scene graphs to extract attributes and relational descriptions. During training, these supervision signals are aligned in a shared latent space via contrastive objectives, allowing the model to distill structured effect knowledge into a compact representation. At test time, the learned effect representation is fused with the action segment features, providing enriched context for downstream mistake detection.

While action effects provide valuable cues, effective mistake detection also requires modeling the execution itself. To this end, we develop a prompt-based detector that evaluates whether an action segment aligns with its execution process within the task context. Rather than relying on frame-level classifiers, our detector encodes each action using a learnable representation and aligns it with a task-specific textual prompt in a contrastive manner. By modeling both the temporal dynamics and the resulting effect of an action, the model captures rich semantic cues to distinguish subtle execution errors as well as outcome discrepancies. Our approach is evaluated on two challenging egocentric datasets, EgoPER \citep{lee2024error} and CaptainCook4D \citep{peddi2023captaincook4d}, achieving state-of-the-art performance. Extensive ablations further demonstrate the necessity and effectiveness of action effect modeling in improving mistake detection.

In summary, our contributions are threefold:
\begin{itemize}[leftmargin=0.5cm]
\vspace{-2mm}
    \item We formulate mistake detection as a probabilistic marginalization over latent action effects, decomposing the task into effect frame sampling, effect modeling, and mistake classification.
    \item We propose Action Effect Modeling to learn effect-aware action representations  by capturing object state and spatial relationships using complementary visual and symbolic cues.
    \item We develop a prompt-based detector that aligns action segments with task-specific textual prompts, enabling effective detection of both execution errors and outcome errors in context.
\end{itemize}

\section{Related Work}

\noindent \textbf{Mistake Detection} consists of two main branches: online detection \citep{jang2019epic, ding2023every, narasimhan2023learning, ashutosh2024video, flaborea2024prego, plini2024ti, soran2015generating, jang2023multimodal} and offline detection \citep{sener2022assembly101, wang2023holoassist, haneji2024egooops, peddi2023captaincook4d, schoonbeek2024industreal, ghoddoosian2023weakly, lee2024error, mazzamuto2024eyes, storks2024explainable, huang2025modeling}. Online mistake detection focuses on detecting sequential mistakes in real time, while offline detection aims to recognize incorrect action execution. This work primarily addresses offline mistake detection.

Zooming into offline detection, early approaches relied on training binary classifiers with visual or multimodal input (\eg, text, audio, depth) to learn implicit action representations, without explicitly modeling the underlying action dynamics \citep{sener2022assembly101, haneji2024egooops, peddi2023captaincook4d}. More recently, \citet{lee2024error} introduced a prototype-based method that explicitly models actions by clustering features during training to learn representative action embeddings. \citet{huang2025modeling} utilized task graphs to predict possible next steps and reconstructed action feature in the latent space. Both methods realized error detection through comparison of similarity between the predicted and input action features at inference time. 

While these methods achieve strong performance, they predominantly focus on modeling the execution process. However, the inherent variability of human behavior poses a significant challenge for detecting fine-grained mistakes solely from execution dynamics. This limitation motivates us to go beyond modeling the action itself by incorporating action effects as complementary signals, thereby enhancing the robustness and reliability of mistake detection.

\noindent \textbf{Video Anomaly Detection} \citep{wu2019deep, gong2019memorizing, pu2022audio, wu2022self, al2024coarse, pu2024learning} is closely related to mistake detection, where models are typically trained on normal videos and identify anomalies as statistical deviations, without requiring prior knowledge of specific abnormal events. However, mistake detection in procedural videos is inherently goal-oriented, as it requires an understanding of structured workflows to assess whether an action is executed correctly. In addition, VAD is predominantly used in surveillance systems, where the camera view remains fixed and the dynamics of the scene is relatively stable. In contrast, procedural videos, especially in egocentric views, involve more fine-grained human activities, explicit scene transitions, and dynamic human motion, making action-effect modeling more challenging but crucial for building life assistive system.
\section{Methodology}
\label{sec:method}

\subsection{Problem Formulation} 
Given a procedural video corresponding to a predefined task, our goal is to determine whether each segmented action is executed correctly or contains a mistake. We operate on non-overlapping segments, where each segment is denoted as a tuple $s = (t_s, t_e, a, y)$, with $t_s$ and $t_e$ denoting the start and end timestamps, $a$ the action label, and $y \in \{0, 1\}$ the binary mistake label ($y = 1$ indicates a mistake). Following the one-class classification (OCC) setting \citep{lee2024error,huang2025modeling}, the training set includes only normal actions ($y = 0$), while the test set contains both correct and erroneous segments. 

Formally, the objective is to estimate the probability of a mistake given a visual representation of the segment, expressed as $P(\hat{y} \mid \mathbf{X})$, where $\mathbf{X}$ denotes the encoded segment feature and $\hat{y}$ is the predicted mistake label. Unlike prior methods that classify segments solely based on execution dynamics, we argue that procedural correctness depends on both the execution process and its outcome. This motivates a probabilistic formulation in which mistake detection is expressed as a marginalization over effect frames and the latent effect descriptors:

\begin{equation}
\label{eq:all_discrete}
\sum_{i=1}^{K} \; \sum_{f_e}
\underbrace{P\left(\hat{y} \mid \mathbf{X}, e_{i}\right)}_{\text{mistake detection}} \cdot 
\underbrace{P\left(e_{i} \mid f_e, \mathbf{X}\right)}_{\text{effect-aware learning}} \cdot
\underbrace{P\left(f_e \mid \mathbf{X}\right)}_{\text{frame sampling}} ,
\end{equation}
where $f_e$ denotes the discrete effect frame sampled from the video, and $\mathbf{e}_i$ represents the $i$-th possible descriptor of the action effect extracted from $f_e$. $K$ is the total number of effect descriptors, which in our case includes two complementary perspectives: object states and spatial relationships, as they reveal common mistakes in procedural activities. Here we assume that once the segment feature $\mathbf{X}$ and the effect descriptor $e_i$ are obtained, the specific effect frame $f_e$ does not provide additional information for mistake prediction, i.e., $\hat y \perp f_e \mid (\mathbf X, e_i)$, and $P(\hat{y}\mid \mathbf{X}, e_i, f_e) = P(\hat{y}\mid \mathbf{X}, e_i)$. Intuitively, Eq. \ref{eq:all_discrete} indicates that we consider every possible effect frame and each of its associated action-effect descriptors, estimate the likelihood of each $(f_e, e_i)$ combination, evaluate whether it suggests a mistake, and then aggregate these weighted probabilities to obtain the final prediction.

This formulation provides a structured view of the mistake detection pipeline: frame sampling, effect-aware learning, and mistake classification, bridging the gap between how an action is performed and what outcomes it produces. Next, we will build a framework following this formulation.

\subsection{Framework Overview}

\begin{wrapfigure}[16]{r}{0.4\textwidth}
    \vspace{-12mm}
    \centering
    \includegraphics[width=\linewidth]{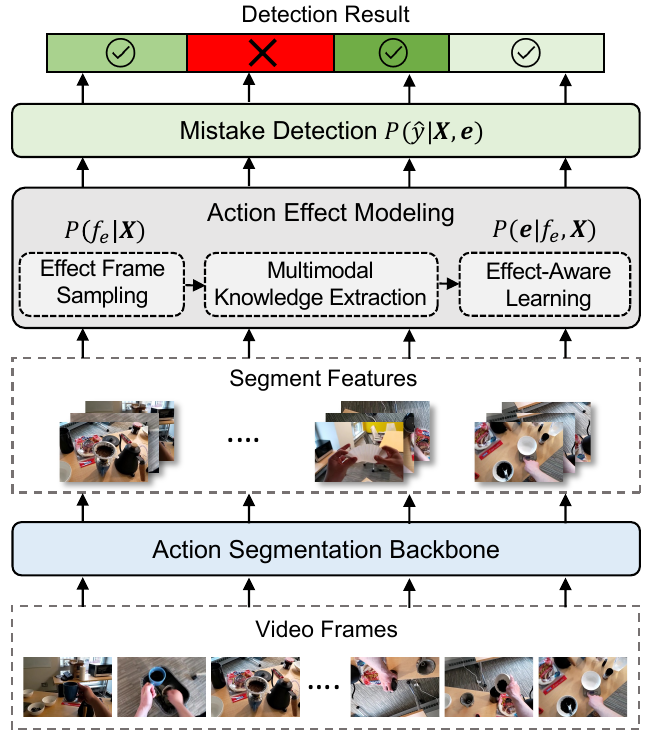}
    \vspace{-6mm}
    \caption{Framework overview.}
    % \vspace{20pts}
    \label{fig:framework}
\end{wrapfigure}

Figure~\ref{fig:framework} illustrates the overall framework. We first adopt an action segmentation backbone to obtain frame-wise features and non-overlapping action segments. The frame features are concatenated along the temporal dimension based on segment boundaries. The resulting segment features $\mathbf{X}$ are fed into the Action Effect Modeling (AEM) module which (i) performs effect frame sampling $P(f_e \mid \mathbf{X})$ to identify the frame most indicative of the outcome, (ii) extracts multimodal effect knowledge $P(\mathbf{e}\mid f_e, \mathbf{X})$ including object states and spatial relationships, and (iii) learns an effect-aware representation aligned across visual and symbolic cues. Finally, the enriched segment features are passed into the mistake detection module, which models $P(\hat{y} \mid \mathbf{X}, \mathbf{e})$ using a prompt-based detector. In the following, we detail three core components: action segmentation, effect modeling and mistake detection. 

\subsection{Action Segmentation}
Following \citet{lee2024error} and \citet{huang2025modeling}, we adopt ActionFormer \citep{zhang2022actionformer} as the action segmentation backbone, which outputs a feature pyramid with multi-scale temporal resolutions. Inspired by prior work \citep{yang2024dyfadet,meng2020ar}, we design a dynamic fusion module to refine temporal representations by adaptively aggregating the multi-scale features from the backbone, while preserving the hierarchical pyramid structure for segment boundary regression. This design aims to provide more discriminative segment features $\mathbf{X}$ for subsequent action effect modeling and mistake detection. Please refer to Appendix \ref{section:a1} for details about this module.

\subsection{Action Effect Modeling}
\begin{figure}[!t]
    \centering
    \includegraphics[width=1.0\linewidth]{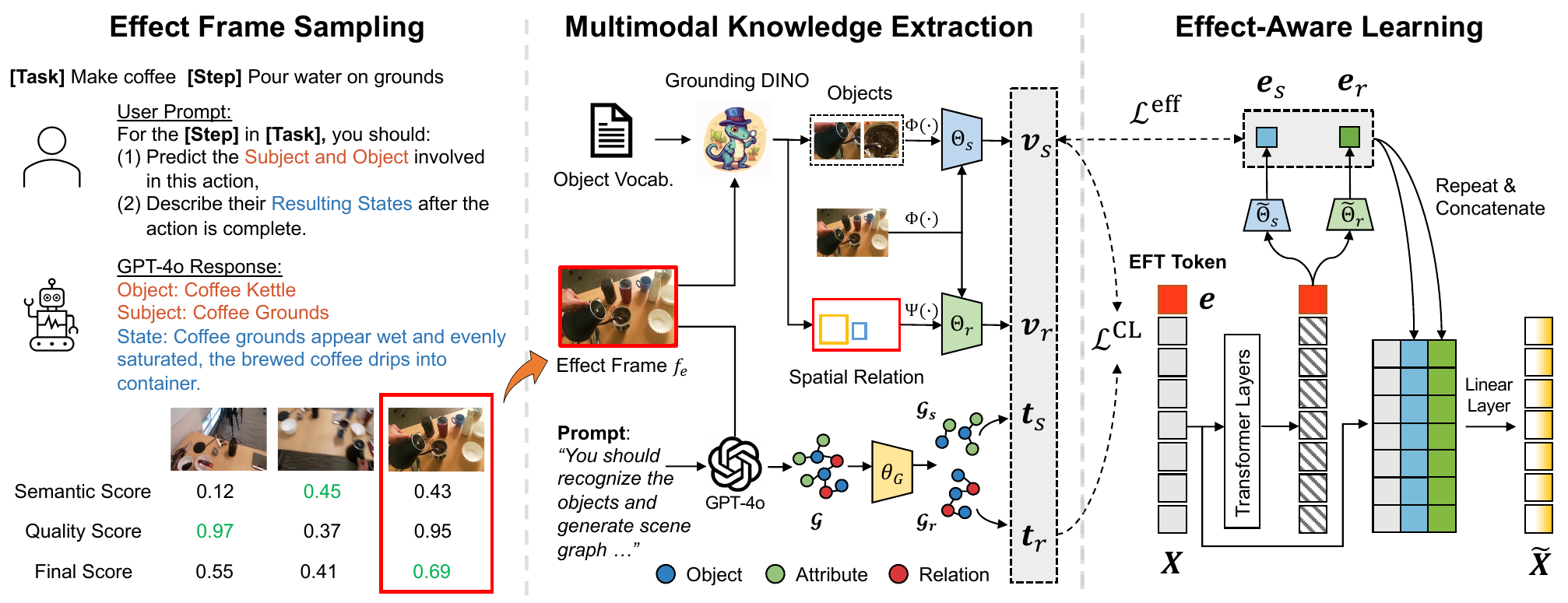}
    \vspace{-12pt}
    \caption{Overview of the Action Effect Modeling (AEM) module. Effect frame sampling and multimodal knowledge extraction are only used to learn effect state and relation projectors $\Theta_s, \Theta_r$ during training. Both of them are skipped during testing.}
    \vspace{-10pt}
    \label{fig:effect_model}
\end{figure}

While the segment feature encodes the temporal dynamics of action execution, it is insufficient to capture the subtle yet critical cues manifested in the action’s outcome. To bridge this gap, AEM integrates action effects by extracting multimodal features from selected effect frames and using them as external supervision. These features guide a learnable \emph{effect token}, which distills outcome semantics into the action representation.  As illustrated in Figure~\ref{fig:effect_model}, the entire AEM pipeline consists of three steps: effect frame sampling, multimodal knowledge extraction, and effect-aware learning. It produces enriched effect-aware action representations that capture both execution and outcome.

\textbf{Effect Frame Sampling.}
To identify the frame that best captures the outcome of a given action, we rank candidate frames within each action segment based on semantic relevance and visual clarity. For semantic relevance, motivated by \citet{niu2024schema}, we use GPT-4o \citep{hurst2024gpt} to generate textual descriptions of the anticipated post-action states. Then we compute the frame-wise similarity between the segment’s visual feature $\mathbf{X}$ and the description embeddings obtained from a pre-trained text encoder, followed by softmax normalization. For visual clarity, we apply the Laplacian operator to estimate the sharpness of each frame within the segment and normalize the scores between the $5th$ and $95th$ percentiles. The final ranking score combines both semantic relevance and visual clarity, and the top-ranked frame is selected as the effect frame.

\textbf{Multimodal Knowledge Extraction.}
From the selected effect frame, we extract cues capturing both object states and spatial configurations, as they can reveal errors that occur in most procedural activities. We adopt a dual-branch strategy: a visual branch for grounded object-centric features, and a textual branch for symbolic scene abstraction.

In the visual branch, we use Grounding DINO \citep{liu2024grounding} to detect relevant objects, then we extract object state and relation features $\mathbf{v}_{s},\mathbf{v}_{r}$ by concatenating and projecting embeddings:
\begin{equation}
\mathbf{v}_{s} = \Theta_s \left( \Phi(f_e) \,\|\, (\Phi(f^{o_i}_e)\big\|_{i=1}^N) \right), \quad
\mathbf{v}_{r} = \Theta_r \left( \Phi(f_e) \,\|\, (\Psi(p^{o_i}_e)\big\|_{i=1}^N) \right),
\end{equation}
where $f_e$ is the effect frame, $f^{o_i}_e$ is the cropped $i$-th object regions, and $p^{o_i}_e$ is the corresponding coordinates of the object bounding boxes. $\Phi(\cdot)$ is a pre-trained image encoder, $\Psi(\cdot)$ is a positional encoding layer, and $\Theta_s, \Theta_r$ are two different multilayer perceptrons (MLPs). $\|$ denotes channel-wise concatenation, and $(\cdot)\big\|_{i=1}^N$ represents the successive concatenation of multiple embeddings.

Although visual grounding provides object-centric features that capture appearance and spatial cues from the effect frame, it lacks high-level semantic abstraction and relational reasoning. To address this, we introduce a textual scene graph branch to model the symbolic structure of the action effect. Given an effect frame and objects detected by Grounding DINO, we leverage the recognition and reasoning capabilities of a Multimodal Large Language Model (MLLM) by prompting GPT-4o to generate a scene graph $\mathcal{G} = (\mathcal{V}, \mathcal{E})$, where $\mathcal{V}$ contains three types of nodes: \emph{object}, \emph{relation}, and \emph{attribute}, and $\mathcal{E}$ represents the set of directed edges. We encode the textual labels of all nodes in $\mathcal{V}$ using a pre-trained text encoder to obtain node embeddings, which are then fed into a Graph Neural Network (GNN) \citep{velivckovic2017graph} to compute context-aware features $\mathbf{t}_v$ for each node.

To abstract finer-grained effect information from text modality, we propose to decompose the generated scene graph $\mathcal{G}$ into two disentangled subgraphs: a state subgraph $\mathcal{G}_{s} = (\mathcal{V}_{s}, \mathcal{E}_{s})$ and a relation subgraph $\mathcal{G}_{r} = (\mathcal{V}_{r}, \mathcal{E}_{r})$. Specifically, the state subgraph includes the object nodes along with their adjacent attribute nodes, capturing object-level changes such as color or texture. The relation subgraph consists of the same object nodes and the intermediate relation node connecting them, representing spatial relationships such as “above” or “inside.”Given $\mathbf{t}_v$ as the GNN feature of each node, we compute the textual features for object states and spatial relations by applying the following average pooling over subgraphs, where $|\mathcal{V}_{s}|$ and $|\mathcal{V}_{r}|$ are the number of nodes in each subgraph:
\begin{equation}
\mathbf{t}_{s} = \frac{\sum_{v \in \mathcal{V}_{s}} \mathbf{t}_{v}}{|\mathcal{V}_{s}|}, \,
\mathbf{t}_{r} = \frac{\sum_{v \in \mathcal{V}_{r}} \mathbf{t}_{v}}{|\mathcal{V}_{r}|},
\end{equation}

\textbf{Effect-Aware Learning.}
After acquiring multimodal effect features, a central challenge is how to effectively integrate these features into the action representation. A naive approach is to concatenate the effect and action segment features, but this would require querying MLLMs during inference, introducing prohibitive computational overhead.  To circumvent this, we treat the effect features as supervision signals and employ a distillation-inspired strategy. Specifically, we introduce a learnable effect token that implicitly captures task-specific action effects through self-attention. During training, this token is aligned with the multimodal effect features to distill external knowledge into a compact representation. Crucially, at inference time the framework relies only on the learned token, eliminating any dependency on external models to ensure efficient deployment.

As shown in Figure \ref{fig:effect_model}, we append the learnable effect token $\mathbf{e}$ to the segment-level features, which are then fed into Transformer encoding layers for temporal modeling. Subsequently, we align the projected effect tokens with the multimodal effect features in the state and relation latent spaces, respectively, which is formulated as:
\begin{equation}
\begin{aligned}
  \mathcal{L}^\text{eff}_s &= ||\widetilde{\Theta}_s(\mathbf{e}) - \mathbf{v}_{s} ||^{2} + ||\widetilde{\Theta}_s(\mathbf{e}) - \mathbf{t}_{s} ||^{2}, \\
  \mathcal{L}^\text{eff}_r &= ||\widetilde{\Theta}_r(\mathbf{e}) - \mathbf{v}_{r} ||^{2} + ||\widetilde{\Theta}_r(\mathbf{e}) - \mathbf{t}_{r} ||^{2},
\end{aligned}
\end{equation}
where \( \widetilde{\Theta}_s(\cdot) \) and \( \widetilde{\Theta}_r(\cdot) \) are spatial and relation projection layers, respectively. This enables efficient learning of a compact and generalizable effect-aware representation without relying on expensive effect-frame computation by MLLMs during inference time.

To enhance the consistency between the two supervision signals, we introduce a contrastive objective that explicitly aligns their visual and textual representations. While the effect loss encourages the learnable token to match each modality individually, this objective ensures that both modalities encode the same underlying semantics from complementary perspectives. Taking the object state as an example, we treat the visual and textual features \((\mathbf{v}_s, \mathbf{t}_{s})\) as positive pairs, and all other samples in the batch as negatives. The contrastive loss is formulated as:
\begin{equation}
    \label{eq: v2g_loss}
    \mathcal{L}^\text{CL}_s = \sum_{i} \left[ -\log \frac{\exp\left(\text{cos}(\mathbf{v}_s^{i}, \mathbf{t}_s^{i}) / \rho\right)}{\sum_{j} \exp\left(\text{cos}(\mathbf{v}_s^{i}, \mathbf{t}_s^{j}) / \rho\right)} \right],
\end{equation}
where \(i\) and \(j\) index samples in the batch, \(\text{cos}(\cdot)\) denotes the cosine similarity, and \(\rho\) is a temperature coefficient. Similarly, we can calculate the relation contrastive loss as \(\mathcal{L}^\text{CL}_r\). 
Together, these objectives enable the effect token to learn a compact multimodal representation of action effects, which is further integrated into the segment features as:
\begin{equation}
    \widetilde{\mathbf{X}}=\mathcal{F}(\mathbf{X} \,\|\, \widetilde{\Theta}_s(\mathbf{e}) \,\|\, \widetilde{\Theta}_r(\mathbf{e}))
\end{equation}
where $\|$ denotes channel-wise concatenation and $\mathcal{F}(\cdot)$ is a linear projection layer. The resulting fused segment features serve as a strong signal for downstream mistake detection.

\subsection{Mistake Detection}
\label{sec:detector}
Following previous work \citep{lee2024error, huang2025modeling}, we adopt an one-class classification (OCC) setting for mistake detection, which learns canonical action patterns from correct samples only. Given the effect-enhanced segment features, we first perform an average-pooling over the enhanced effect-aware segment feature $\widetilde{\mathbf{X}}$ across temporal dimension to obtain the action embedding $\mathbf{x}_a$. Instead of learning a binary classifier, we introduce a prompt-based detector to increase task-specific discriminability. For each action label $a$, we construct a template-based textual prompt $\mathcal{P}_a$ (e.g., “An image showing \texttt{[ACTION]} for \texttt{[TASK]}”), and obtain its representation by:
\begin{equation}
    \mathbf{y}_a = E\left( \mathbf{p} \,\|\, \text{Embed}(\text{Tokenize}(\mathcal{P}_a)) \right),
\end{equation}
where $\mathbf{p}$ is a learnable prefix embedding and $E(\cdot)$ is the frozen text encoder. $\|$ denotes concatenation.

During training, we align the action embedding with its corresponding prompt features while repulsing it from other prompt embeddings through a contrastive objective:
\begin{equation}
\mathcal{L}^{\text{det}} = -\frac{1}{B} \sum_{i=1}^{B} \log \left( \frac{
\exp\left( \text{cos}(\mathbf{x}_a^i, \mathbf{y}_{a}^{i}) / \rho \right)}{
\sum_{c=1}^{C} \exp\left( \text{cos}(\mathbf{x}_a^i, \mathbf{y}_{a}^{c}) / \rho \right)
} \right),
\label{eq:cls}
\end{equation}
where $\mathbf{x}^i_a$ denotes the $i$-th action embedding in a mini-batch $B$, and $C$ is the total number of actions in a specific task. 
During inference, we compute the similarity score $sim(\mathbf{x}_{a}, \mathbf{y}_{a})$ between the action embedding and its corresponding textual prompt by $sim(\mathbf{x}_{a}, \mathbf{y}_{a})=\text{cos}(\mathbf{x}_{a}, \mathbf{y}_{a}) / \rho$. The final prediction of mistake probability can be obtained by $P(\hat{y} \mid \mathbf{X}, \mathbf{e})=1-sim(\mathbf{x}_{a}, \mathbf{y}_{a})$. The mistake detection result is determined by thresholding the mistake probability:
\begin{equation}
\hat{y} =
\begin{cases}
1, & P(\hat{y} \mid \mathbf{X}, \mathbf{e}) > \tau, \\
0, & \text{otherwise}.
\end{cases}
\label{eq:error_def}
\end{equation}
where $\hat{y}$ is the predicted mistake label and $\tau$ is a predefined threshold.

\subsection{Training Objective}
Finally, we jointly optimize three objectives to train the model in an end-to-end fashion: the segmentation loss $\mathcal{L}^{\text{seg}}$ following \citet{zhang2022actionformer}, the action-effect modeling loss $\mathcal{L}^{\text{eff}}$ and $\mathcal{L}^{\text{CL}}$, and the mistake detection loss $\mathcal{L}^{\text{det}}$. The overall training objective is defined as:
\begin{equation}
    \mathcal{L}=\mathcal{L}^{\text{seg}}+\mathcal{L}^{\text{eff}}+\mathcal{L}^{\text{CL}}+\mathcal{L}^{\text{det}},
\end{equation}
where $\mathcal{L}^{\text{eff}}$ and $\mathcal{L}^{\text{CL}}$ comprise both state and relation terms. By incorporating action-effect modeling into the training process, our framework learns to attend to semantically meaningful object state transitions and spatial relationships, thereby enhancing both representation and mistake detection.
\section{Experiments}
\label{sec:exp}
\definecolor{darkgray}{gray}{0.5}

\renewcommand{\arraystretch}{1.0}
\begin{table*}[!t]
\centering
\footnotesize
\caption{Mistake detection results on the EgoPER dataset \citep{lee2024error}. AUC and EDA are in percentage (\%). Best results are \textbf{bold}, and second-best are \underline{underlined}.}
% \vspace{-4pt}
\resizebox{\textwidth}{!}{
\begin{tabular}{@{}lcccccccccccc@{}}
\toprule
\multirow{2}{*}{Method} & \multicolumn{2}{c}{Quesadilla} & \multicolumn{2}{c}{Oatmeal} & \multicolumn{2}{c}{Pinwheel} & \multicolumn{2}{c}{Coffee} & \multicolumn{2}{c}{Tea} & \multicolumn{2}{c}{All} \\
& AUC & EDA & AUC & EDA & AUC & EDA & AUC & EDA & AUC & EDA & AUC & EDA \\
\midrule
Random & 50.0 & 19.9 & 50.0 & 11.8 & 50.0 & 15.7 & 50.0 & 8.2 & 50.0 & 17.0 & 50.0 & 14.5 \\
HF\textsuperscript{2}-VAD & 62.6 & 34.5 & 62.3 & 25.4 & 52.7 & 29.1 & 59.6 & 10.0 & 62.1 & 36.6 & 59.9 & 27.1 \\
HF\textsuperscript{2}-VAD + SSPCAB & 60.9 & 30.4 & 61.9 & 25.3 & 51.7 & 33.9 & 60.1 & 10.0 & 63.2 & 35.4 & 59.6 & 27.0 \\
S3R & 51.8 & 52.6 & 61.6 & 47.8 & 52.4 & 50.5 & 51.0 & 16.3 & 57.9 & 47.8 & 54.9 & 43.0 \\
EgoPED & 65.6 & \underline{62.7} & 65.1 & 51.4 & 55.0 & 59.6 & 58.3 & 55.3 & \underline{66.0} & 56.0 & 62.0 & 57.0 \\
AMNAR & \underline{71.9} & 61.4 & \underline{75.4} & \underline{65.0} & \underline{65.4} & \textbf{65.0} & \underline{67.8} & \textbf{73.5} & 61.9 & \underline{57.0} & \underline{68.5} & \underline{64.4} \\
\noalign{\vskip 1mm}
\cdashline{1-13}
\noalign{\vskip 1mm}
\textbf{Ours} & \textbf{80.8} & \textbf{68.1} & \textbf{77.0} & \textbf{68.6} & \textbf{69.9} & \underline{61.2} & \textbf{70.3} & \underline{66.4} & \textbf{71.1} & \textbf{69.4} & \textbf{73.8} & \textbf{66.7} \\
\midrule
\bottomrule
\end{tabular}
}
% \vspace{-4pt}
\label{tab:comparison}
\end{table*}

\begin{table}[!t]
\centering
% \scriptsize
\footnotesize
\caption{Mistake detection results on the CaptainCook4D dataset
\citep{peddi2023captaincook4d}.}
% \vspace{-4pt}
% \begin{tabular*}{0.6\linewidth}{@{\extracolsep{\fill}}lccc@{}}
\begin{tabular}{lccc}
\toprule
Method & Precision & AUC & EDA \\
\midrule
Random & 44.9 & 51.2 & 49.7 \\
EgoPED & 56.5 & 54.9 & 69.8 \\
AMNAR  & \underline{66.8} & \underline{60.2} & \textbf{72.3} \\
\textbf{Ours} & \textbf{68.1} & \textbf{62.5} & \underline{71.9} \\
\midrule
\bottomrule
\end{tabular}
% \vspace{-6pt}
\label{tab:ccp4d}
\end{table}

\subsection{Experimental Setup}
\noindent \textbf{Mistake Datasets.} We evaluate our method on two egocentric video datasets: EgoPER \citep{lee2024error} and CaptainCook4D \citep{huang2025modeling}. 
The EgoPER dataset consists of 385 egocentric cooking videos spanning 5 recipe categories, with a total duration of 28 hours. Among these, 213 videos contain only correct actions, while the remaining 178 include various execution mistakes. Following \citet{lee2024error}, we use 80\% of the normal videos for training, 10\% for validation, and the remaining 10\% together with all erroneous videos for testing. 
The CaptainCook4D dataset contains 384 videos covering 24 recipes, with a total length of 94.5 hours. Among all recordings, 164 are labeled as normal, while the remaining 220 videos include incorrect action steps. Following \citet{huang2025modeling}, we use all normal recordings for training and the erroneous ones for testing.

\noindent \textbf{Evaluation Metrics.}  We follow prior work \citep{lee2024error, huang2025modeling} by using Area Under the Curve (AUC) and Error Detection Accuracy (EDA)  to jointly measure model's ability to distinguish correct and erroneous actions within each video. AUC assesses frame-level discrimination between errors and non-errors, whereas EDA measures segment-level detection accuracy. Both metrics are computed by aggregating detection results across a densely sampled threshold range from 0 to 1, rather than relying on a single fixed threshold, enabling a more comprehensive evaluation of the model’s behavior across the full spectrum of thresholds.

\noindent \textbf{Implementation Details.} We use off-the-shelf EVA-02 CLIP \citep{eva02} for all image and text encoding. For the EgoPER dataset, we follow \citet{lee2024error} to use additional features based on Active Object Detection (AOD). For effect frame sampling, the final score is an average of the semantic relevance and visual quality scores. Following \citet{lee2024error} and \citet{huang2025modeling}, we independently train and test the model in each recipe and report the average of all recipes as the final performance. Both the effect-frame sampling and multimodal knowledge extraction are performed during the \emph{data pre-processing} stage to avoid additional training overhead. All experiments were conducted with an NVIDIA RTX 6000 Ada GPU (48GB memory). More implementation details can be found in the Appendix \ref{section:a1}.

\subsection{Comparison with State-of-the-art}
We compare our methods with video anomaly detection methods HF\textsuperscript{2}-VAD \citep{liu2021hybrid}, HF\textsuperscript{2}-VAD + SSPCAB \citep{ristea2022self} and S3R \citep{wu2022self}, as well as mistake detection methods EgoPED \citep{lee2024error} and AMNAR \citep{huang2025modeling}. As shown in Table~\ref{tab:comparison}, on the EgoPER dataset, our method outperforms in both metrics on most of the five tasks, with an average improvement of 5.3\% on AUC and 2.3\% on EDA. The results in Table~\ref{tab:ccp4d} further demonstrate its effectiveness on the CaptainCook4D dataset, surpassing AMNAR by 2.7\% in Precision and 2.3\% in AUC. On both datasets, our EDA is occasionally lower than that of AMNAR. This gap is likely due to AMNAR’s use of a dynamic programming block to construct task graphs that mitigate noise in action-segment labels. Since incorporating task graphs lies beyond the scope of our work, we leave it for future exploration.

\begin{table*}[t]
\centering
\renewcommand{\arraystretch}{1.0}
\caption{Ablation studies (a–d), action segmentation results (e), and performance comparison with different scene graph generation models (f) on EgoPER dataset \citep{lee2024error}.}
\vspace{-10pt}

% ---------------- 第一排： (a) | (b)+(c) ----------------
\begin{minipage}{\columnwidth}
\centering
% 左列 (a)
\begin{minipage}{0.5\columnwidth}
\centering
\footnotesize
{\setlength{\tabcolsep}{4pt}
\subcaption{Alignment between multimodal supervision with the projected state and relation effect tokens.}
\label{tab:ablation_a}
\begin{tabular}{ccccccc}
\toprule
\multirow{2}{*}{AEM} & \multicolumn{2}{c}{$\mathcal{L}^\text{eff}_s$} & \multicolumn{2}{c}{$\mathcal{L}^\text{eff}_r$} & \multirow{2}{*}{AUC} & \multirow{2}{*}{EDA} \\
 & Visual & Textual & Visual & Textual & & \\
\midrule
\ding{55} & \multicolumn{4}{c}{\text{N/A}} & 67.6 & 65.6 \\
\midrule
\multirow{6}{*}{\ding{51}}
 & \ding{55} & \ding{55} & \ding{55} & \ding{55} & 67.9 & 65.8 \\
 & \ding{51} & \ding{51} & \ding{55} & \ding{55} & 68.4 & 66.1 \\
 & \ding{55} & \ding{55} & \ding{51} & \ding{51} & 69.4 & 66.3 \\
 & \ding{51} & \ding{55} & \ding{51} & \ding{55} & \underline{71.7} & \underline{66.4} \\
 & \ding{55} & \ding{51} & \ding{55} & \ding{51} & 69.9 & 66.0 \\
 & \ding{51} & \ding{51} & \ding{51} & \ding{51} & \textbf{73.8} & \textbf{66.7} \\
\midrule
\bottomrule
\end{tabular}}
\end{minipage}%
\hfill
% \vspace{-6pt}
% 右列： (b) 和 (c) 竖向堆叠，二者各自居中
\begin{minipage}{0.44\columnwidth}
\centering
\footnotesize
{\setlength{\tabcolsep}{12pt}
\vspace{12pt}
\subcaption{Effect-frame sampling strategies.}
\label{tab:ablation_b}
\begin{tabular}{lcc}
\toprule
Method & AUC & EDA \\
\midrule
w/o Effect & 67.6 & 65.6 \\
Last Frame & \underline{70.6} & \underline{65.7} \\
\textbf{Ours} & \textbf{73.8} & \textbf{66.7} \\
\midrule
\bottomrule
\end{tabular}}

\vspace{0.5em}

\footnotesize
{\setlength{\tabcolsep}{13pt}
\vspace{6pt}
\subcaption{Dynamic cross-level interactions.}
\label{tab:ablation_c}
\begin{tabular}{lcc}
\toprule
Method & AUC & EDA \\
\midrule
AMNAR & 68.5 & \underline{64.4} \\
w/o Dyn & \underline{71.8} & 63.4 \\
\textbf{w/ Dyn} & \textbf{73.8} & \textbf{66.7} \\
\midrule
\bottomrule
\end{tabular}}
\end{minipage}

\end{minipage}

\vspace{0.5em}

% ---------------- 第二排： (d) | (e) ----------------
\begin{minipage}{\columnwidth}
\centering

% 左列 (d)
\hspace*{0.05cm}
\begin{minipage}{0.56\columnwidth}
\centering
\footnotesize
{\setlength{\tabcolsep}{12pt}
\vspace{-3pt}
\subcaption{Alignment of supervisions in state and relation spaces.}
\label{tab:ablation_d}
\begin{tabular}{cccc}
\toprule
$\mathcal{L}^{\text{CL}}_s$ & $\mathcal{L}^{\text{CL}}_r$ & AUC & EDA \\
\midrule
\ding{55} & \ding{55} & 66.8 & 64.7 \\
\ding{51} & \ding{55} & 69.9 & 64.5 \\
\ding{55} & \ding{51} & \underline{72.6} & \underline{65.1} \\
\ding{51} & \ding{51} & \textbf{73.8} & \textbf{66.7} \\
\midrule
\bottomrule
\end{tabular}}
\end{minipage}%
\hfill
% 右列 (e)
\begin{minipage}{0.43\columnwidth}
\centering
\footnotesize
{\setlength{\tabcolsep}{6pt}
\vspace{7pt}
\subcaption{Action segmentation performance.}
\label{tab:ablation_e}
\begin{tabular}{lcccc}
\toprule
Method & IoU & Edit & F1@0.5 & Acc \\
\midrule
EgoPED  & 44.6 & 61.3 & 47.5 & 68.5 \\
AMNAR   & \underline{56.3} & \underline{69.4} & \underline{57.3} & \textbf{75.3} \\
\textbf{Ours} & \textbf{58.5} & \textbf{69.7} & \textbf{58.5} & \underline{73.5} \\
\midrule
\bottomrule
\end{tabular}}
\end{minipage}
\end{minipage}

\vspace{1.2em}

% ---------------- 第三排：蓝色大表 ----------------
\begingroup
\centering
\footnotesize
\subcaption{Performance comparison of our method with different scene graph generation models.}
% \vspace{-4pt}

\resizebox{\textwidth}{!}{
\begin{tabular}{@{}lcccccccccccc@{}}
\toprule
\multirow{2}{*}{Models} &
\multicolumn{2}{c}{Quesadilla} &
\multicolumn{2}{c}{Oatmeal} &
\multicolumn{2}{c}{Pinwheel} &
\multicolumn{2}{c}{Coffee} &
\multicolumn{2}{c}{Tea} &
\multicolumn{2}{c}{All} \\
& AUC & EDA & AUC & EDA & AUC & EDA & AUC & EDA & AUC & EDA & AUC & EDA \\
\midrule
Qwen3-VL 
& 77.8 & \textbf{69.0} 
& \textbf{77.3} & \textbf{68.8}
& \textbf{70.0} & \textbf{62.2}
& 70.2 & 64.3
& \textbf{71.4} & 68.7
& 73.3 & 66.6 \\

GPT-4o 
& \textbf{80.8} & 68.1
& 77.0 & 68.6
& 69.9 & 61.2
& \textbf{70.3} & \textbf{66.4}
& 71.1 & \textbf{69.4}
& \textbf{73.8} & \textbf{66.7} \\
\midrule
\bottomrule
\end{tabular}
}
\label{tab:opensource_results}
\endgroup

\vspace{-4pt}
\label{tab:ablation}
\end{table*}

\renewcommand{\arraystretch}{1.0}

\subsection{Ablation Studies}
In the ablation study, we aim to answer the following questions:
\begin{enumerate}[leftmargin=0.5cm, itemsep=0.7pt, topsep=1pt, parsep=0pt]
\vspace{-2mm}
    \item Is Action Effect Modeling effective and generalizable?
    \item Which is more important: object state effect or spatial relation effect?
    \item How do visual and textual signals contribute to AEM?
    \item Is aligning external multimodal features necessary?
    \item How do different effect-frame sampling strategies perform?
    \item Does multi-scale dynamic fusion module contribute to mistake detection?
    \item Can open-source MLLMs replace the expensive closed-source models?
\end{enumerate}

\emph{1) Action Effect Modeling.}
As shown in Table~\ref{tab:ablation_a}, the first row reports the performance of baseline, which already achieves a comparable performance to AMNAR. This is attributed to the prompt-based detector’s ability to capture task-specific action representations, thereby enhancing the discriminability of action execution. The $2^{\text{nd}}$ row further demonstrates that the learnable token can implicitly model action effects without relying on contextual knowledge provided by Grounding DINO or GPT-4o. After introducing different forms of external supervision, our model achieves consistent improvements, verifying its generalizability. Finally, when combining all multimodal effect supervision signals, our model achieves the best performance due to supervised effect modeling.

\emph{2) State vs. Relation Effect.}
The $3^{\text{rd}}$ and $4^{\text{th}}$ rows of Table~\ref{tab:ablation_a} show the impact of state and relational effects. Compared to implicit token learning, explicitly introducing supervision signals for state or relational effects leads to performance gains, where relational supervision performs better, yielding a 1.0\% improvement in AUC over state supervision. We argue that spatial relationships between objects provide more consistent and discriminative cues for action correctness. In contrast, object state changes are often subtle and visually ambiguous, making them more difficult to capture reliably.

\emph{3) Visual vs. Textual Supervision.}
The $5^{\text{th}}$ and $6^{\text{th}}$ rows of Table~\ref{tab:ablation_a} report the impact of incorporating visual and textual features as external knowledge. Compared to baseline, using visual features improves AUC to 71.7\% and EDA to 66.4\%, while textual features yield a smaller gain. We argue that visual features are more effective because they directly capture object appearance and spatial layout, offering fine-grained cues to reflect execution-related visual differences. In contrast, textual features, although semantically informative, are abstracted from generated scene graphs and may introduce noise or miss subtle visual distinctions in complex egocentric environments.

\emph{4) Effect Feature Alignment.}
We further study the impact of aligning multimodal supervision features in state and relation effect representation spaces. As shown in Table~\ref{tab:ablation_d}, using unaligned visual and textual supervisions leads to a suboptimal AUC of 66.8\% and EDA of 64.7\%, even lower than the model without effect modeling, highlighting the importance of cross-modal alignment. Introducing contrastive alignment in either the state space or the relation space improves AUC to 69.9\% and 72.6\%, improves EDA to 64.5\% and 65.1\%. The stronger gain from relation alignment suggests that object spatial relationships play a more critical role in modeling action effects. Finally, aligning both state and relation representations achieves the best AUC, demonstrating the complementary nature of these two effect types and the effectiveness of joint cross-modal supervision.

\emph{5) Effect Frame Sampling.}
To assess the effectiveness of our effect-frame sampling strategy, we compare it with a naive baseline that selects the last frame of each action segment, assuming that action effects are most visible at the end. As shown in Table~\ref{tab:ablation_b}, this heuristic raises AUC to 70.6\%, confirming that incorporating effect information benefits mistake detection. In contrast, our proposed strategy, which jointly considers semantic relevance and visual clarity further, boosts AUC to 73.8\% and EDA to 66.7\%. These results imply that a more informed selection of effect frames yields more discriminative representations, thereby improving detection performance.

\emph{6) Dynamic Fusion Module.}
Table~\ref{tab:ablation_c} evaluates whether enhancing multi-scale temporal representations contributes to the final detection performance. The results show that incorporating dynamic fusion yields a 2\% improvement in AUC and a 3.3\% gain in EDA, highlighting the importance of expressive action segment representations for downstream mistake detection. Notably, even without this module, our model still surpasses AMNAR by 3.3\% AUC and achieves comparable EDA, further highlighting the superiority of our overall framework.

\emph{7) Alternative open-source MLLM.}
In this work, we employ a state-of-the-art closed-source MLLM GPT-4o \citep{hurst2024gpt} to generate action scene graphs that provide supervision signals for our method. To assess whether alternative models can fulfill the same role, we replace GPT-4o with the latest open-source MLLM Qwen3-VL (30B) \citep{Qwen3-VL}. As shown in Table~\ref{tab:opensource_results}, the scene graphs generated by Qwen3-VL yield mistake detection performance comparable to that by GPT-4o, indicating that open-source models can serve as a cost-effective alternative for the generation of scene graphs while maintaining strong performance.

\subsection{Action Segmentation}
Table~\ref{tab:ablation_e} presents a comparison of action segmentation performance. All models, including ours, use ActionFormer \citep{zhang2022actionformer} as the segmentation backbone, ensuring a fair comparison. Our approach outperforms prior methods across most evaluation metrics, highlighting its effectiveness in enhancing action segmentation in addition to mistake detection. These results also indicate that our framework holds promise for broader video understanding tasks where action effects serve as critical cues, such as action recognition \citep{bao2021evidential} and action analysis \citep{huang2025h}.

\subsection{Qualitative Analysis} 
\begin{figure}
    \centering
    \includegraphics[width=1\linewidth]{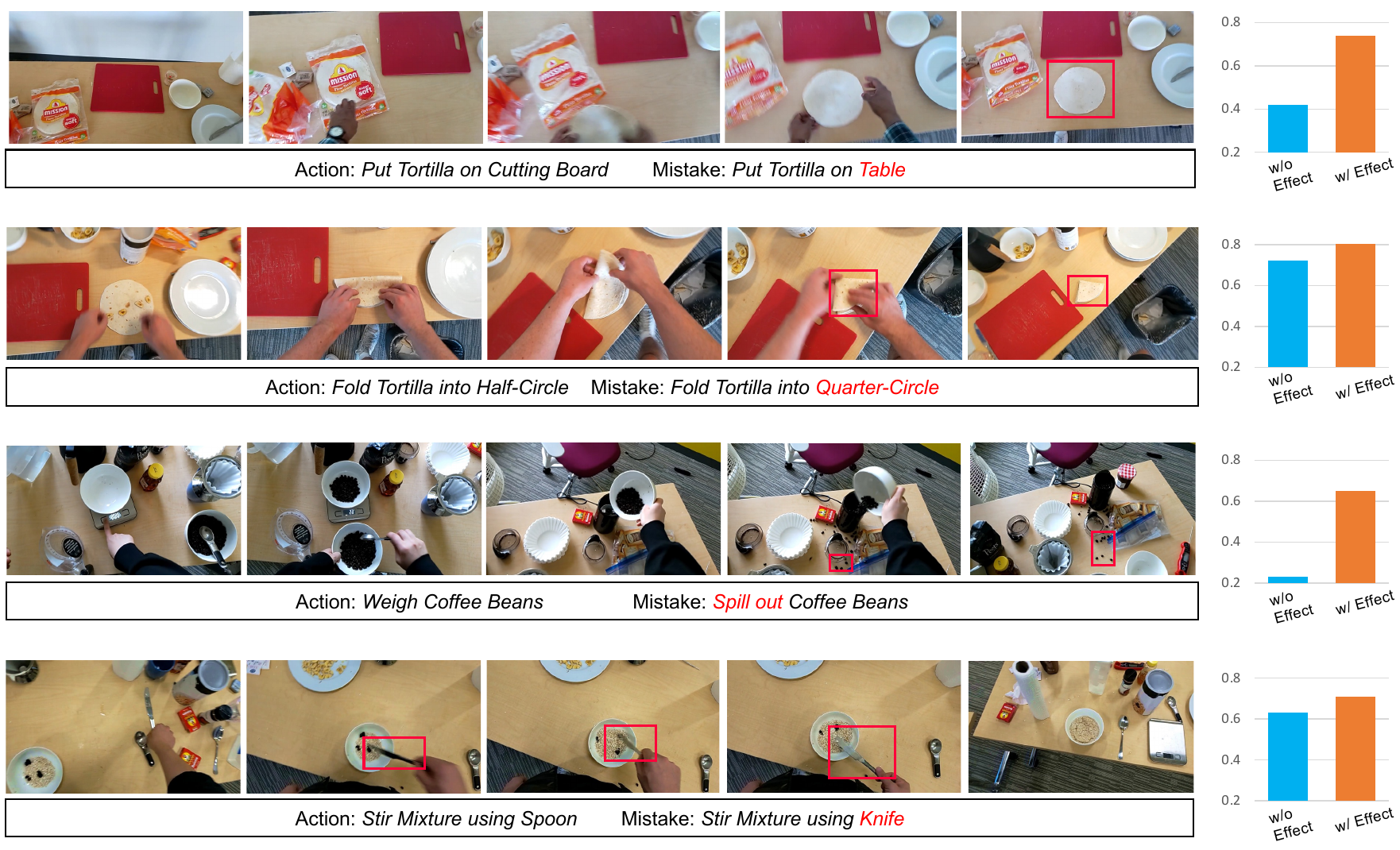}
    \caption{Examples of mistakes occurring in different actions. The right bar charts show  mistake probabilities predicted by models without (in blue) and with (in orange) effect modeling. Red boxes in images are only used to highlight mistake regions for clearer visualization.}
    \label{fig:vis}
\end{figure}

Figure~\ref{fig:vis} shows visualizations of the action segments and the prediction of mistake probabilities. The results highlight two key advantages of our approach: (i) the model effectively detects errors that appear in the final outcomes through action–effect modeling (row 1 to row 3), and (ii) it can also identify execution errors that arise during the process enabled by the prompt-based detector, even when the visual outcome seems correct (row 4). Together, these findings demonstrate the complementarity between effect-aware representations and temporal execution modeling in capturing a broader spectrum of procedural errors. More visualization can be found in the Appendix \ref{section:a3}.

\section{Conclusion}
We presented a unified framework for procedural mistake detection that explicitly models the relationship between action execution and its resulting effect. The proposed Action Effect Modeling (AEM) enriches action representations with effect-aware cues by integrating object states and spatial relationships, guided through multimodal supervision. Combined with a prompt-based detector, our framework effectively captures both subtle execution errors and outcome discrepancies, achieving substantial improvements over prior methods on mistake detection benchmarks. Future work includes extending the framework to spatial-temporal action effect modeling for long-range procedural reasoning and improving mistake interpretability via human-understandable explanations generated by Large Language Models (LLMs).
% Future work could extend the approach to multi-step reasoning for more comprehensive procedural activity understanding and mistake detection.

\section*{Acknowledgment}
This research was partially sponsored by the MSU Jenison Fund and Army Research Office (under Grant Number W911NF-24-1-0385). The views and conclusions contained in this document are those of the authors and should not be interpreted as representing the official policies, either expressed or implied, of the Army Research Office or the U.S. Government. The U.S. Government is authorized to reproduce and distribute reprints for Government purposes notwithstanding any copyright notation herein. The authors thank Wentao Bao for insightful suggestions on the methodological aspects of this work.

\bibliography{iclr2026_conference}

@String(ECCV= {Eur. Conf. Comput. Vis.})

@String(ECCV  = {ECCV})

@inproceedings{jang2019epic,
  title={Epic-tent: An egocentric video dataset for camping tent assembly},
  author={Jang, Youngkyoon and Sullivan, Brian and Ludwig, Casimir and Gilchrist, Iain and Damen, Dima and Mayol-Cuevas, Walterio},
  booktitle={Proceedings of the IEEE/CVF International Conference on Computer Vision Workshops},
  pages={0--0},
  year={2019}
}

@inproceedings{sener2022assembly101,
  title={Assembly101: A large-scale multi-view video dataset for understanding procedural activities},
  author={Sener, Fadime and Chatterjee, Dibyadip and Shelepov, Daniel and He, Kun and Singhania, Dipika and Wang, Robert and Yao, Angela},
  booktitle={Proceedings of the IEEE/CVF Conference on Computer Vision and Pattern Recognition},
  pages={21096--21106},
  year={2022}
}

@inproceedings{wang2023holoassist,
  title={Holoassist: an egocentric human interaction dataset for interactive ai assistants in the real world},
  author={Wang, Xin and Kwon, Taein and Rad, Mahdi and Pan, Bowen and Chakraborty, Ishani and Andrist, Sean and Bohus, Dan and Feniello, Ashley and Tekin, Bugra and Frujeri, Felipe Vieira and others},
  booktitle={Proceedings of the IEEE/CVF International Conference on Computer Vision},
  pages={20270--20281},
  year={2023}
}

@article{haneji2024egooops,
  title={EgoOops: A Dataset for Mistake Action Detection from Egocentric Videos with Procedural Texts},
  author={Haneji, Yuto and Nishimura, Taichi and Kameko, Hirotaka and Shirai, Keisuke and Yoshida, Tomoya and Kajimura, Keiya and Yamamoto, Koki and Cui, Taiyu and Nishimoto, Tomohiro and Mori, Shinsuke},
  journal={arXiv preprint arXiv:2410.05343},
  year={2024}
}

@article{peddi2023captaincook4d,
  title={CaptainCook4D: A dataset for understanding errors in procedural activities},
  author={Peddi, Rohith and Arya, Shivvrat and Challa, Bharath and Pallapothula, Likhitha and Vyas, Akshay and Wang, Jikai and Zhang, Qifan and Komaragiri, Vasundhara and Ragan, Eric and Ruozzi, Nicholas and others},
  journal={arXiv preprint arXiv:2312.14556},
  year={2023}
}

@inproceedings{schoonbeek2024industreal,
  title={Industreal: A dataset for procedure step recognition handling execution errors in egocentric videos in an industrial-like setting},
  author={Schoonbeek, Tim J and Houben, Tim and Onvlee, Hans and Van der Sommen, Fons and others},
  booktitle={Proceedings of the IEEE/CVF Winter Conference on Applications of Computer Vision},
  pages={4365--4374},
  year={2024}
}

@inproceedings{ghoddoosian2023weakly,
  title={Weakly-supervised action segmentation and unseen error detection in anomalous instructional videos},
  author={Ghoddoosian, Reza and Dwivedi, Isht and Agarwal, Nakul and Dariush, Behzad},
  booktitle={Proceedings of the IEEE/CVF International Conference on Computer Vision},
  pages={10128--10138},
  year={2023}
}

@inproceedings{lee2024error,
  title={Error detection in egocentric procedural task videos},
  author={Lee, Shih-Po and Lu, Zijia and Zhang, Zekun and Hoai, Minh and Elhamifar, Ehsan},
  booktitle={Proceedings of the IEEE/CVF Conference on Computer Vision and Pattern Recognition},
  pages={18655--18666},
  year={2024}
}

@article{mazzamuto2024eyes,
  title={Eyes Wide Unshut: Unsupervised Mistake Detection in Egocentric Video by Detecting Unpredictable Gaze},
  author={Mazzamuto, Michele and Furnari, Antonino and Farinella, Giovanni Maria},
  journal={arXiv preprint arXiv:2406.08379},
  year={2024}
}

@article{storks2024explainable,
  title={Explainable Procedural Mistake Detection},
  author={Storks, Shane and Bar-Yossef, Itamar and Li, Yayuan and Zhang, Zheyuan and Corso, Jason J and Chai, Joyce},
  journal={arXiv preprint arXiv:2412.11927},
  year={2024}
}

@article{ding2023every,
  title={Every mistake counts in assembly},
  author={Ding, Guodong and Sener, Fadime and Ma, Shugao and Yao, Angela},
  journal={arXiv preprint arXiv:2307.16453},
  year={2023}
}

@article{narasimhan2023learning,
  title={Learning and verification of task structure in instructional videos},
  author={Narasimhan, Medhini and Yu, Licheng and Bell, Sean and Zhang, Ning and Darrell, Trevor},
  journal={arXiv preprint arXiv:2303.13519},
  year={2023}
}

@article{jang2023multimodal,
  title={Multimodal subtask graph generation from instructional videos},
  author={Jang, Yunseok and Sohn, Sungryull and Logeswaran, Lajanugen and Luo, Tiange and Lee, Moontae and Lee, Honglak},
  journal={arXiv preprint arXiv:2302.08672},
  year={2023}
}

@inproceedings{flaborea2024prego,
  title={PREGO: online mistake detection in PRocedural EGOcentric videos},
  author={Flaborea, Alessandro and di Melendugno, Guido Maria D'Amely and Plini, Leonardo and Scofano, Luca and De Matteis, Edoardo and Furnari, Antonino and Farinella, Giovanni Maria and Galasso, Fabio},
  booktitle={Proceedings of the IEEE/CVF Conference on Computer Vision and Pattern Recognition},
  pages={18483--18492},
  year={2024}
}

@article{plini2024ti,
  title={TI-PREGO: Chain of Thought and In-Context Learning for Online Mistake Detection in PRocedural EGOcentric Videos},
  author={Plini, Leonardo and Scofano, Luca and De Matteis, Edoardo and di Melendugno, Guido Maria D'Amely and Flaborea, Alessandro and Sanchietti, Andrea and Farinella, Giovanni Maria and Galasso, Fabio and Furnari, Antonino},
  journal={arXiv preprint arXiv:2411.02570},
  year={2024}
}

@inproceedings{soran2015generating,
  title={Generating notifications for missing actions: Don't forget to turn the lights off!},
  author={Soran, Bilge and Farhadi, Ali and Shapiro, Linda},
  booktitle={Proceedings of the IEEE International Conference on Computer Vision},
  pages={4669--4677},
  year={2015}
}

@article{niu2024schema,
  title={SCHEMA: State CHangEs MAtter for Procedure Planning in Instructional Videos},
  author={Niu, Yulei and Guo, Wenliang and Chen, Long and Lin, Xudong and Chang, Shih-Fu},
  journal={arXiv preprint arXiv:2403.01599},
  year={2024}
}

@inproceedings{liu2021hybrid,
  title={A hybrid video anomaly detection framework via memory-augmented flow reconstruction and flow-guided frame prediction},
  author={Liu, Zhian and Nie, Yongwei and Long, Chengjiang and Zhang, Qing and Li, Guiqing},
  booktitle={Proceedings of the IEEE/CVF international conference on computer vision},
  pages={13588--13597},
  year={2021}
}

@inproceedings{ristea2022self,
  title={Self-supervised predictive convolutional attentive block for anomaly detection},
  author={Ristea, Nicolae-C{\u{a}}t{\u{a}}lin and Madan, Neelu and Ionescu, Radu Tudor and Nasrollahi, Kamal and Khan, Fahad Shahbaz and Moeslund, Thomas B and Shah, Mubarak},
  booktitle={Proceedings of the IEEE/CVF conference on computer vision and pattern recognition},
  pages={13576--13586},
  year={2022}
}

@inproceedings{wu2022self,
  title={Self-supervised sparse representation for video anomaly detection},
  author={Wu, Jhih-Ciang and Hsieh, He-Yen and Chen, Ding-Jie and Fuh, Chiou-Shann and Liu, Tyng-Luh},
  booktitle={European Conference on Computer Vision},
  pages={729--745},
  year={2022},
  organization={Springer}
}

@article{ashutosh2024video,
  title={Video-mined task graphs for keystep recognition in instructional videos},
  author={Ashutosh, Kumar and Ramakrishnan, Santhosh Kumar and Afouras, Triantafyllos and Grauman, Kristen},
  journal={Advances in Neural Information Processing Systems},
  volume={36},
  year={2024}
}

@inproceedings{pu2022audio,
  title={Audio-guided attention network for weakly supervised violence detection},
  author={Pu, Yujiang and Wu, Xiaoyu},
  booktitle={2022 2nd International Conference on Consumer Electronics and Computer Engineering (ICCECE)},
  pages={219--223},
  year={2022},
  organization={IEEE}
}

@article{pu2024learning,
  title={Learning prompt-enhanced context features for weakly-supervised video anomaly detection},
  author={Pu, Yujiang and Wu, Xiaoyu and Yang, Lulu and Wang, Shengjin},
  journal={IEEE Transactions on Image Processing},
  year={2024},
  publisher={IEEE}
}

@inproceedings{zhang2022actionformer,
  title={Actionformer: Localizing moments of actions with transformers},
  author={Zhang, Chen-Lin and Wu, Jianxin and Li, Yin},
  booktitle={European Conference on Computer Vision},
  pages={492--510},
  year={2022},
  organization={Springer}
}

@article{wu2019deep,
  title={A deep one-class neural network for anomalous event detection in complex scenes},
  author={Wu, Peng and Liu, Jing and Shen, Fang},
  journal={IEEE transactions on neural networks and learning systems},
  volume={31},
  number={7},
  pages={2609--2622},
  year={2019},
  publisher={IEEE}
}

@inproceedings{gong2019memorizing,
  title={Memorizing normality to detect anomaly: Memory-augmented deep autoencoder for unsupervised anomaly detection},
  author={Gong, Dong and Liu, Lingqiao and Le, Vuong and Saha, Budhaditya and Mansour, Moussa Reda and Venkatesh, Svetha and Hengel, Anton van den},
  booktitle={Proceedings of the IEEE/CVF international conference on computer vision},
  pages={1705--1714},
  year={2019}
}

@inproceedings{al2024coarse,
  title={A coarse-to-fine pseudo-labeling (c2fpl) framework for unsupervised video anomaly detection},
  author={Al-Lahham, Anas and Tastan, Nurbek and Zaheer, Muhammad Zaigham and Nandakumar, Karthik},
  booktitle={Proceedings of the IEEE/CVF Winter Conference on Applications of Computer Vision},
  pages={6793--6802},
  year={2024}
}

@article{velivckovic2017graph,
  title={Graph attention networks},
  author={Veli{\v{c}}kovi{\'c}, Petar and Cucurull, Guillem and Casanova, Arantxa and Romero, Adriana and Lio, Pietro and Bengio, Yoshua},
  journal={arXiv preprint arXiv:1710.10903},
  year={2017}
}

@inproceedings{liu2024grounding,
  title={Grounding dino: Marrying dino with grounded pre-training for open-set object detection},
  author={Liu, Shilong and Zeng, Zhaoyang and Ren, Tianhe and Li, Feng and Zhang, Hao and Yang, Jie and Jiang, Qing and Li, Chunyuan and Yang, Jianwei and Su, Hang and others},
  booktitle={European Conference on Computer Vision},
  pages={38--55},
  year={2024},
  organization={Springer}
}

@article{hurst2024gpt,
  title={Gpt-4o system card},
  author={Hurst, Aaron and Lerer, Adam and Goucher, Adam P and Perelman, Adam and Ramesh, Aditya and Clark, Aidan and Ostrow, AJ and Welihinda, Akila and Hayes, Alan and Radford, Alec and others},
  journal={arXiv preprint arXiv:2410.21276},
  year={2024}
}

@inproceedings{huang2025modeling,
  title={Modeling Multiple Normal Action Representations for Error Detection in Procedural Tasks},
  author={Huang, Wei-Jin and Li, Yuan-Ming and Xia, Zhi-Wei and Tang, Yu-Ming and Lin, Kun-Yu and Hu, Jian-Fang and Zheng, Wei-Shi},
  booktitle={Proceedings of the Computer Vision and Pattern Recognition Conference},
  pages={27794--27804},
  year={2025}
}

@article{eva02,
  title={Eva-02: A visual representation for neon genesis},
  author={Fang, Yuxin and Sun, Quan and Wang, Xinggang and Huang, Tiejun and Wang, Xinlong and Cao, Yue},
  journal={Image and Vision Computing},
  pages={105171},
  year={2024},
  publisher={Elsevier}
}

@inproceedings{yang2024dyfadet,
  title={DyFADet: Dynamic Feature Aggregation for Temporal Action Detection},
  author={Yang, Le and Zheng, Ziwei and Han, Yizeng and Cheng, Hao and Song, Shiji and Huang, Gao and Li, Fan},
  booktitle={European Conference on Computer Vision (ECCV)},
  year={2024}
}

@inproceedings{meng2020ar,
  title={Ar-net: Adaptive frame resolution for efficient action recognition},
  author={Meng, Yue and Lin, Chung-Ching and Panda, Rameswar and Sattigeri, Prasanna and Karlinsky, Leonid and Oliva, Aude and Saenko, Kate and Feris, Rogerio},
  booktitle={European conference on computer vision},
  pages={86--104},
  year={2020},
  organization={Springer}
}

@inproceedings{bao2021evidential,
  title={Evidential deep learning for open set action recognition},
  author={Bao, Wentao and Yu, Qi and Kong, Yu},
  booktitle={Proceedings of the IEEE/CVF international conference on computer vision},
  pages={13349--13358},
  year={2021}
}

@inproceedings{huang2025h,
  title={H-MoRe: Learning Human-centric Motion Representation for Action Analysis},
  author={Huang, Zhanbo and Liu, Xiaoming and Kong, Yu},
  booktitle={Proceedings of the Computer Vision and Pattern Recognition Conference},
  pages={22702--22713},
  year={2025}
}

@article{li2021prefix,
  title={Prefix-tuning: Optimizing continuous prompts for generation},
  author={Li, Xiang Lisa and Liang, Percy},
  journal={arXiv preprint arXiv:2101.00190},
  year={2021}
}

@misc{openai_gpt5_2025,
  author       = {OpenAI},
  title        = {GPT-5: Large Language Model},
  year         = {2025},
  url          = {https://openai.com/index/introducing-gpt-5/},
  note         = {Accessed: 2025-11-13}
}

@article{Qwen3-VL,
      title={Qwen3-VL Technical Report}, 
      author={Shuai Bai and Yuxuan Cai and Ruizhe Chen and Keqin Chen and Xionghui Chen and Zesen Cheng and Lianghao Deng and Wei Ding and Chang Gao and Chunjiang Ge and Wenbin Ge and Zhifang Guo and Qidong Huang and Jie Huang and Fei Huang and Binyuan Hui and Shutong Jiang and Zhaohai Li and Mingsheng Li and Mei Li and Kaixin Li and Zicheng Lin and Junyang Lin and Xuejing Liu and Jiawei Liu and Chenglong Liu and Yang Liu and Dayiheng Liu and Shixuan Liu and Dunjie Lu and Ruilin Luo and Chenxu Lv and Rui Men and Lingchen Meng and Xuancheng Ren and Xingzhang Ren and Sibo Song and Yuchong Sun and Jun Tang and Jianhong Tu and Jianqiang Wan and Peng Wang and Pengfei Wang and Qiuyue Wang and Yuxuan Wang and Tianbao Xie and Yiheng Xu and Haiyang Xu and Jin Xu and Zhibo Yang and Mingkun Yang and Jianxin Yang and An Yang and Bowen Yu and Fei Zhang and Hang Zhang and Xi Zhang and Bo Zheng and Humen Zhong and Jingren Zhou and Fan Zhou and Jing Zhou and Yuanzhi Zhu and Ke Zhu},
	  journal={arXiv preprint arXiv:2511.21631},
      year={2025}
}
\bibliographystyle{iclr2026_conference}

\newpage

\appendix
\section*{Appendix}
This appendix is organized as follows:
\begin{itemize}[leftmargin=0.5cm]
\vspace{-2mm}
    \item \textbf{Appendix \ref{section:a1}} introduces more details about the implementation.
    \item \textbf{Appendix \ref{section:a2}} presents more experimental results and analysis.
    \item \textbf{Appendix \ref{section:a3}} provides more visualizations for demonstration.
    \item \textbf{Appendix \ref{section:a4}} shares further discussion on limitations and future directions.
    \item \textbf{Appendix \ref{section:a5}} clarifies the use of the large language model.
\end{itemize}
\section{implementation Details}
\label{section:a1}
\subsection{Dynamic Fusion Module}
\begin{figure}[h]
    \centering
    \includegraphics[width=1\linewidth]{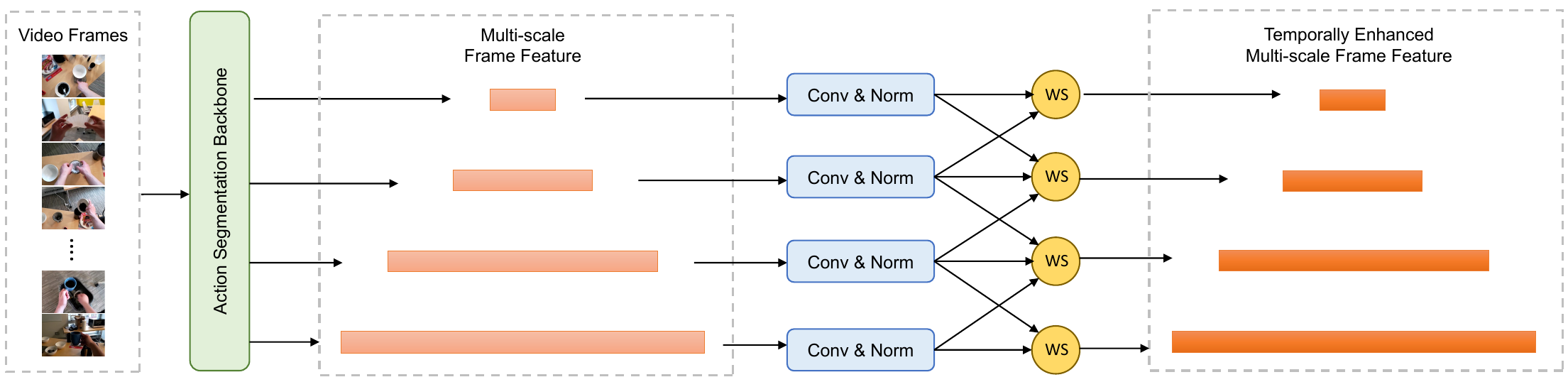}
    \caption{Structure of the dynamic fusion module. Each \emph{Conv\&Norm} block represents a convolutional layer followed by a normalization layer, and each \emph{WS} block denotes a weighted-sum operation.}
    \label{fig:dyn_fucion}
\end{figure}

As described in Section~\ref{sec:method}, we design a simple yet effective dynamic fusion module to bridge interactions among multi-scale frame features produced by the ActionFormer backbone, thereby refining temporal representations. The internal structure of this module is illustrated in Figure~\ref{fig:dyn_fucion}. Frame features at each level, with different temporal lengths, are first passed through a convolutional layer followed by a normalization layer to capture temporal information at varying granularities. For each feature, the features from its adjacent levels are up-sampled or down-sampled along the temporal dimension via linear interpolation to match its length, after which a weighted sum is applied. The weights are learnable and specific to each feature. The resulting features preserve their original length and dimensionality while being enriched with aggregated temporal information. An ablation study in Table~\ref{tab:ablation_e} validates the effectiveness of this dynamic fusion module for mistake detection.

\subsection{Effect-frame Sampling}
In this work, we perform action-effect modeling by leveraging the effect knowledge embedded in effect frames. To sample effect frames within action segments, we first predict key objects that indicate the action effect and then apply semantic scoring to evaluate the relevance of each video frame to the effect. Figure~\ref{fig:gen_desc} presents the complete GPT-4o prompt used for key-object prediction and generation of textual effect descriptions.

\subsection{Action Scene Graph}
\label{sec:graph}
In this work, we construct a symbolic action scene graph from sampled effect frames to generate textual supervision for effect learning. This subsection details the implementation of the action scene graph, including: (1) generating the scene graph by prompting GPT-4o and parsing its output, (2) decomposing the graph into object-state and spatial-relation subgraphs, and (3) presenting a concrete example that illustrates the overall process.

\subsubsection{Scene Graph Construction}
\begin{figure}[h]
    \centering
    \begin{subfigure}{\linewidth}
        \centering
        \includegraphics[width=\linewidth]{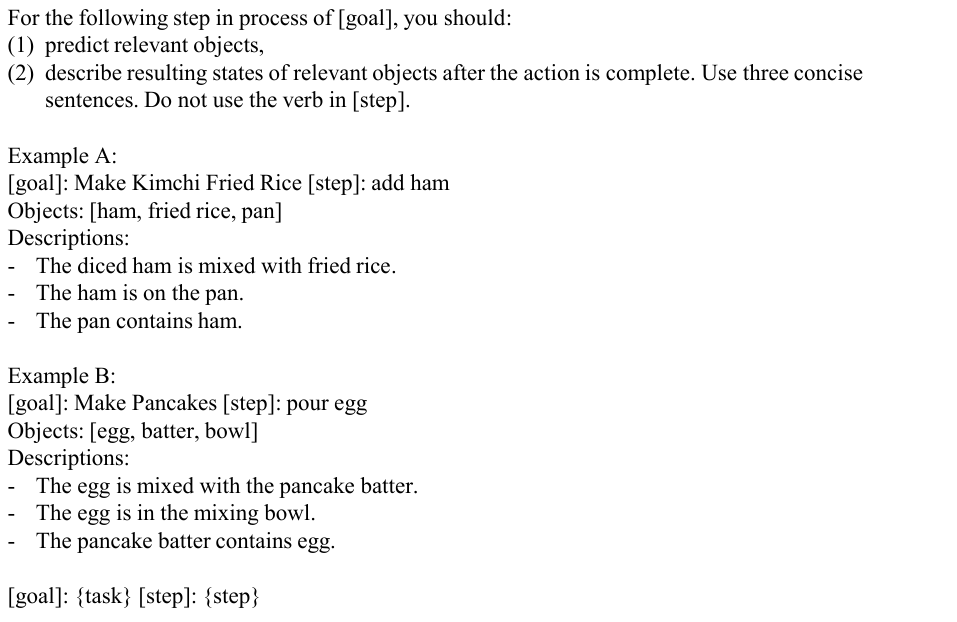}
        \vspace{-6mm}
        \caption{Prompt used to generate relevant objects and resulting outcome descriptions.}
        % \vspace{-6pt}
        \label{fig:gen_desc}
    \end{subfigure}

    \begin{subfigure}{\linewidth}
        \centering
        \includegraphics[width=\linewidth]{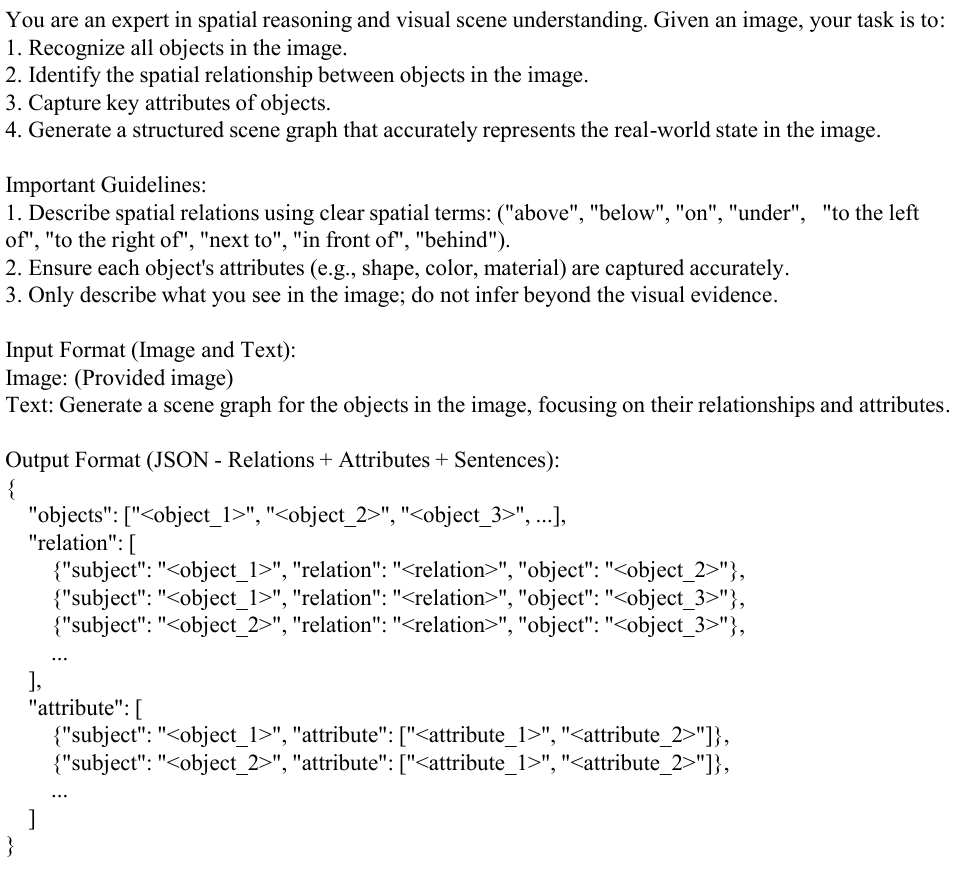}
        \vspace{-6mm}
        \caption{Prompt used to recognize object attributes and spatial relationships.}
        \label{fig:gen_pmt}
    \end{subfigure}
    
    \caption{Prompts used in this work.}
    \label{fig:prompts}
\end{figure}

We prompt the powerful GPT-4o to recognize the action scene and output the scene information. Figure \ref{fig:gen_pmt} shows the complete prompt we used. Scene information is returned in the format of the dictionary (JSON file), and we design algorithms to parse the output and construct the action scene graph based on it.  Algorithm \ref{alg:scene_graph_construct} describes the graph construction process, where the input includes the object relation dictionary (\textit{Relation\_Dict}) and the object attribute dictionary (\textit{Attribute\_Dict}) generated by the VLM. The output consists of two dictionaries: one containing the graph nodes categorized by type (\textit{Nodes}) and the other storing the edges connecting pairs of nodes (\textit{Nodes}).

\subsubsection{Action Scene Graph Decomposition}
Algorithm \ref{alg:scene_graph_decompose} describes the graph decomposition process, where the input includes \textit{Nodes} and \textit{Nodes} obtained from graph construction, as well as the specified action-related objects generated by effect frame sampling. The outputs are separate sub-graphs of object state (\textit{S\text{-}Graph}) and object spatial relationships (\textit{R\text{-}Graph}).

\subsubsection{Example of Scene Graph Processing}
\begin{figure}[h]
    \centering
    \includegraphics[width=0.75\linewidth]{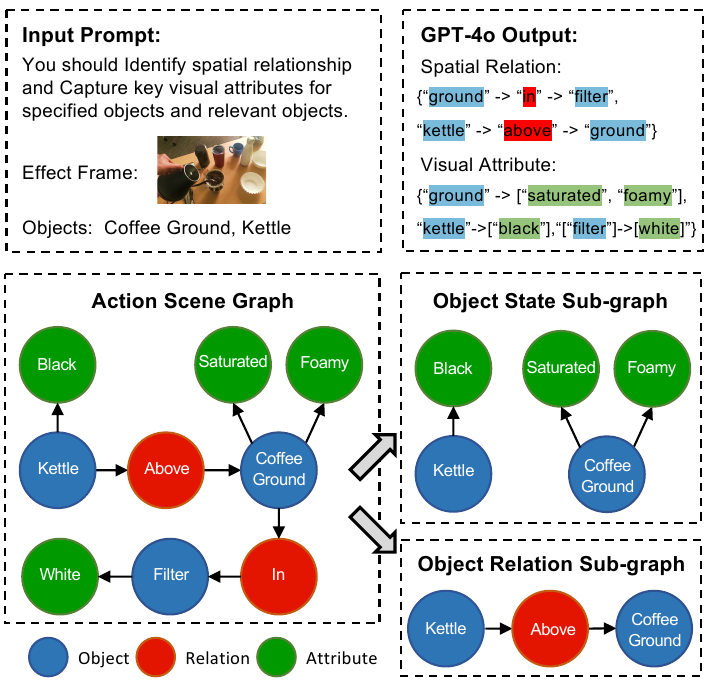}
    \caption{An example showing the construction and decomposition of action scene graph.}
    \label{fig:graph_example}
\end{figure}

Figure \ref{fig:graph_example} illustrates how an action scene graph is constructed and decomposed for the effect frame of \emph{Pouring Water on Coffee Ground} within the task of \emph{Making Coffee}.

\subsection{Solution to Failed Knowledge Extraction}
During effect-frame preprocessing for multimodal feature extraction, several failure cases were observed. Grounding DINO may miss relevant objects or yield low-confidence detections, while GPT-4o may fail to recognize the image, leading to missing object states or spatial relations. To address these issues, we introduce an Effect Mask. The mask is assigned a value of $1$ only when both Grounding DINO and GPT-4o successfully process the effect frame; otherwise, it is set to $0$. During feature alignment, only instances with a mask value of 1 contribute to loss computation and are used to generate enhanced segment features for mistake detection. For masked-out instances, effect modeling is bypassed, and segment features from the segmentation backbone are directly fed into the mistake detector.

\begin{algorithm}[!t]
\caption{Action Scene Graph Construction}
\label{alg:scene_graph_construct}
\renewcommand{\algorithmicrequire}{\textbf{Input:}}
\renewcommand{\algorithmicensure}{\textbf{Output:}}
\resizebox{\columnwidth}{!}{% <-- 自动缩放到一栏宽
\begin{minipage}{\columnwidth}
% \scriptsize  % <-- 缩小字体
\begin{algorithmic}[1]
    \Require \textit{Relation\_Dict}, \textit{Attribute\_Dict}
    \Ensure \textit{Nodes}, \textit{Edges}

    \Statex \textit{// Initialization.}
    \State \textit{Nodes} $\gets$ \text{dict({"Object":[], "Attribute":[], "Relation":[]})}  
    \State \textit{Edges} $\gets$ \text{list([])}

    \vspace{6pt}
    \Statex \textit{// Add object-relation edges.}
    \For{each $(obj_A, rel, obj_B)$ in \textit{Relation\_Dict}}
        \If {$obj_A \notin \textit{Nodes}["Object"]$}
            \State Append $obj_A$ to \textit{Nodes}["Object"]
        \EndIf
        \If {$obj_B \notin \textit{Nodes}["Object"]$}
            \State Append $obj_B$ to \textit{Nodes}["Object"]
        \EndIf
        \If {$rel \notin \textit{Nodes}["Relation"]$}
            \State Append $rel$ to \textit{Nodes}["Relation"]
        \EndIf
        \State Append $(obj_A, rel)$, $(rel, obj_B)$ to \textit{Edges}
    \EndFor

    \vspace{6pt}
    \Statex \textit{// Add object-attribute edges}
    \For{each $(obj, attr)$ in \textit{Attribute\_Dict}}
        \If {$obj \notin \textit{Nodes}["Object"]$}
            \State Append $obj$ to \textit{Nodes}["Object"]
        \EndIf
        \If {$attr \notin \textit{Nodes}["Attribute"]$}
            \State Append $attr$ to \textit{Nodes}["Attribute"]
        \EndIf
        \State Append $(obj, attr)$ to \textit{Edges}
    \EndFor

    \vspace{6pt}
    \State \Return \textit{Nodes}, \textit{Edges}
\end{algorithmic}
\end{minipage}
} % <-- 结束缩放
\end{algorithm}

\begin{algorithm}[!t]
\caption{Action Scene Graph Decomposition}
\label{alg:scene_graph_decompose}
\renewcommand{\algorithmicrequire}{\textbf{Input:}}
\renewcommand{\algorithmicensure}{\textbf{Output:}}
\begin{algorithmic}[1]
    \Require Extracted graph $Nodes$, graph $Edges$, and key objects $Objs = \{obj_i\}_{i=1}^{N}$
    \Ensure State graph $S\text{-}Graph$, relation graph $R\text{-}Graph$

    \Statex \textit{// Initialization.}
    \State Initialize $S\text{-}Graph \gets$ dict(\{``Nodes'':[], ``Edges'':[]\})
    \State Initialize $R\text{-}Graph \gets$ dict(\{``Nodes'':[], ``Edges'':[]\})

    \vspace{6pt}
    \Statex \textit{// Construct the state graph from object-attribute edges}
    \For{each edge $(a,b)$ in $Edges$}
        \If{$a \in Objs$ \textbf{and} $b \in Nodes["Attribute"]$}
            \State Append $a$ and $b$ to $S\text{-}Graph["Nodes"]$
            \State Append $(a,b)$ to $S\text{-}Graph["Edges"]$
        \EndIf
    \EndFor

    \vspace{6pt}
    \Statex \textit{// Construct the state graph from object-attribute edges}
    \For{each edge $(a,b)$ in $Edges$}
        \If{$a \in Objs$ \textbf{and} $b \in Nodes["Relation"]$}
            \State Append $a$ and $b$ to $R\text{-}Graph["Nodes"]$
            \State Append $(a,b)$ to $R\text{-}Graph["Edges"]$
        \EndIf
    \EndFor

    \vspace{6pt}
    \State \Return $S\text{-}Graph, R\text{-}Graph$
\end{algorithmic}
\end{algorithm}

\clearpage
\section{Experiments}
\label{section:a2}

\subsection{Impact of VLM on performance}
\begin{table}[h]
\centering
% \footnotesize
\caption{Performance by using different VLMs for action scene generation on EgoPER dataset.}
\renewcommand{\arraystretch}{1.0}
\setlength{\tabcolsep}{4pt} % 控制列间距
\begin{tabular*}{\linewidth}{@{\extracolsep{\fill}} lcccccccccccc}
\toprule
\multirow{2}{*}{Model} &
\multicolumn{2}{c}{Quesadilla} &
\multicolumn{2}{c}{Oatmeal} &
\multicolumn{2}{c}{Pinwheel} &
\multicolumn{2}{c}{Coffee} &
\multicolumn{2}{c}{Tea} &
\multicolumn{2}{c}{All} \\
& AUC & EDA & AUC & EDA & AUC & EDA & AUC & EDA & AUC & EDA & AUC & EDA \\
\midrule
Qwen3-VL  & 77.8 & 69.0 & 77.3 & 68.8 & 70.0 & \textbf{62.2} & 70.2 & 64.3 & \textbf{71.4} & 68.7 & 73.3 & 66.6 \\
GPT-4o  & \textbf{80.8} & 68.1 & 77.0 & 68.6 & 69.9 & 61.2 & 70.3 & 66.4 & 71.1 & 69.4 & 73.8 & 66.7 \\
GPT-5          & 78.9 & \textbf{69.1} & \textbf{79.1} & \textbf{70.1} & \textbf{70.4} & 61.9 & \textbf{70.4} & \textbf{66.4} & 71.3 & \textbf{69.8} & \textbf{74.0} & \textbf{67.5} \\
\midrule
\bottomrule
\end{tabular*}
\label{tab:ablation_vlm}
\end{table}

At the time of completing the main paper, we employed the SOTA closed-source VLM GPT-4o \citep{hurst2024gpt} and the open-source VLM Qwen3-VL \citep{Qwen3-VL} to generate action scene graphs. In this subsection of the Appendix, we extend Table \ref{tab:opensource_results} in the main paper by providing further evaluation results of our proposed method using a more recent GPT-5 model \citep{openai_gpt5_2025}. Table \ref{tab:ablation_vlm} shows that the GPT-5 variant achieves the best detection performance which exceeds both models, suggesting that our method continues to benefit from ongoing advances in VLMs.

\subsection{Action Segmentation}
\begin{table}[h]
\centering
\caption{Impact of predicted vs. ground-truth action segmentation on EgoPER dataset.}
\setlength{\tabcolsep}{5pt}
\renewcommand{\arraystretch}{1.0}
\setlength{\tabcolsep}{4pt} % 控制列间距
\begin{tabular}{ccccccccccccc}
\toprule
\multirow{2}{*}{Action Seg} & \multicolumn{2}{c}{Quesadilla} & \multicolumn{2}{c}{Oatmeal} & \multicolumn{2}{c}{Pinwheel} & \multicolumn{2}{c}{Coffee} & \multicolumn{2}{c}{Tea} & \multicolumn{2}{c}{Average} \\
& AUC & EDA & AUC & EDA & AUC & EDA & AUC & EDA & AUC & EDA & AUC & EDA \\
\midrule
Pred & 80.8 & 68.1 & 77.0 & 68.6 & 69.9 & 61.2 & 70.3 & 66.4 & 71.1 & 69.4 & 73.8 & 66.7 \\
GT   & 92.8 & 78.4 & 94.2 & 75.0 & 80.7 & 72.0 & 80.8 & 71.3 & 84.8 & 76.7 & 86.7 & 74.7 \\
\midrule
\bottomrule
\end{tabular}
\label{tab:action_seg}
\end{table}

To assess the impact of action segmentation on mistake detection, we evaluated our method by replacing predicted segments with ground-truth segments \textit{during inference} on the test set. As shown in Table \ref{tab:action_seg}, using ground-truth segmentation yields a significant improvement in mistake detection performance, which can be regarded as an upper bound for the current framework. This result suggests that improvements in segmentation can consistently enhance detection performance; therefore, future work will explore leveraging more advanced action segmentation models.

\subsection{Prompt Tuning in Mistake Detector}
\renewcommand{\arraystretch}{1.0} 
\begin{table}[htbp]
\centering
\caption{Ablation results for the learnable prompts on EgoPER dataset.}
\label{tab:ablation_prmpt}
\setlength{\tabcolsep}{4pt}
\begin{tabular}{c cc cc cc cc cc cc}
\toprule
\multirow{2}{*}{Learnable} 
& \multicolumn{2}{c}{Quesadilla} 
& \multicolumn{2}{c}{Oatmeal} 
& \multicolumn{2}{c}{Pinwheel} 
& \multicolumn{2}{c}{Coffee} 
& \multicolumn{2}{c}{Tea} 
& \multicolumn{2}{c}{All} \\
& AUC & EDA & AUC & EDA & AUC & EDA & AUC & EDA & AUC & EDA & AUC & EDA \\
\midrule
\ding{55} & 77.4 & \textbf{68.9} & 76.6 & 57.5 & 69.5 & 60.7 & 68.5 & \textbf{67.0} & \textbf{72.1} & 63.6 & 72.8 & 63.5 \\
\ding{51} & \textbf{80.8} & 68.1 & \textbf{77.0} & \textbf{68.6} & \textbf{69.9} & \textbf{61.2} & \textbf{70.3} & 66.4 & 71.1 & \textbf{69.4} & \textbf{73.8} & \textbf{66.7} \\
\midrule
\bottomrule
\end{tabular}
\end{table}
In the mistake detector, we adopt learnable prompts combined with prefix-tuning techniques \citep{li2021prefix} to capture action patterns. To evaluate the effectiveness of the prompt-based detector, we compare learnable prompts with fixed prompts. Both methods use the same template, “an image showing \texttt{[ACTION]} for \texttt{[TASK]}”, to generate contextual embeddings. The fixed prompt does not include learnable embeddings; instead, it directly uses the frozen text encoder of EVA-02 CLIP \citep{eva02} to extract features. As shown in Table~\ref{tab:ablation_prmpt}, the learnable prompt achieves superior performance on most subtasks as well as on average compared to the fixed prompt. This indicates that fixed textual descriptions, due to the lack of temporal and contextual motion encoding, fail to align effectively with visual action features. Therefore, additional learnable embeddings are necessary to capture the temporal dynamics of actions.

For tasks where the performance of learnable prompts is slightly lower than that of fixed prompts, such as \emph{Make Quesadilla}, we observed that many erroneous actions and their corresponding mistakes are less related to temporal dynamics. For example, \emph{Place tortilla on table} (instead of on cutting board) and \emph{Fold tortilla into a quarter-circle} (instead of a half-circle) already encode sufficient object-related information by using their textual labels. In such cases, our effect-aware model can distinguish the errors effectively, and introducing additional learnable prompts to capture action dynamics brings limited benefit and may even interfere with effect encoding, leading to performance degradation. As this work provides an initial exploration of encoding action temporal dynamics in mistake detection, extending it to a broader range of action labels is an interesting research direction.

\subsection{Effect Frame Scene Recognition}
In this work, we employ GPT-4o \citep{hurst2024gpt} to recognize the environments of effect frames, which are then used to build scene graphs and extract symbolic knowledge. Recognition errors in this process may introduce noise into the effect modeling. Although the main paper demonstrates that effect modeling substantially improves mistake detection, we further conducted a quantitative evaluation of GPT-4o’s recognition performance on complex scenes.

\begin{figure}
    \centering
    \includegraphics[width=\linewidth]{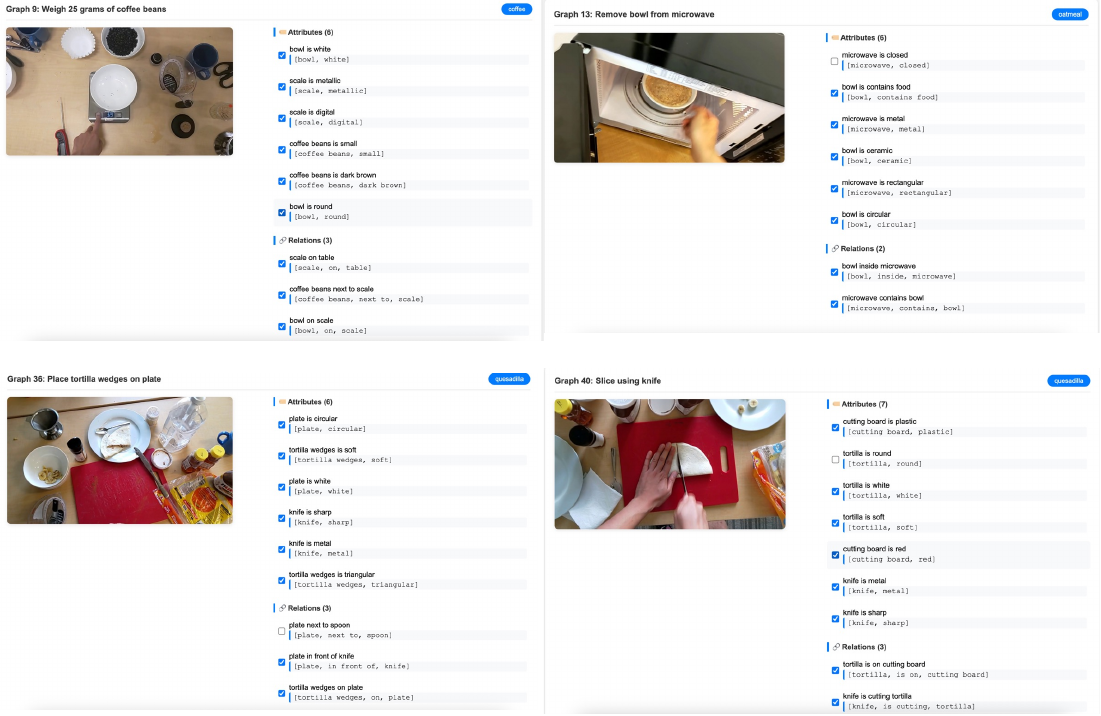}
    \caption{Web interface for scene recognition manual check.}
    \label{fig:interface}
\end{figure}

Specifically, we randomly sampled 20 recognition results from each sub-dataset of the EgoPER dataset \citep{lee2024error}, yielding 100 scene instances in total. As shown in Figure~\ref{fig:interface}, we developed an HTML-based interface to display the recognized attributes and spatial relations for each effect frame. Two graduate students from the computer science department, with no domain conflicts, were recruited and randomly assigned 50 scenes each for manual evaluation. We define two metrics to assess recognition performance: Success Rate, the proportion of effect frames in which all recognition results are correct, and Accuracy, the proportion of correctly recognized items among all recognition results. The final scores were obtained by averaging the results of the two annotators.

Manual evaluation yielded a success rate of 48.4\% and an accuracy of 87.0\%. Although GPT-4o occasionally produces recognition errors, its multi-perspective scene descriptions generally complement one another, enabling reliable reconstruction of object states and spatial relationships. This provides a solid foundation for effect modeling, and we anticipate that such limitations will diminish as multimodal large language models continue to advance.

\clearpage
\section{Visualizations}
\label{section:a3}

\subsection{Action Scene Graph}
\begin{figure}[htbp]
    \centering
    \includegraphics[width=\linewidth]{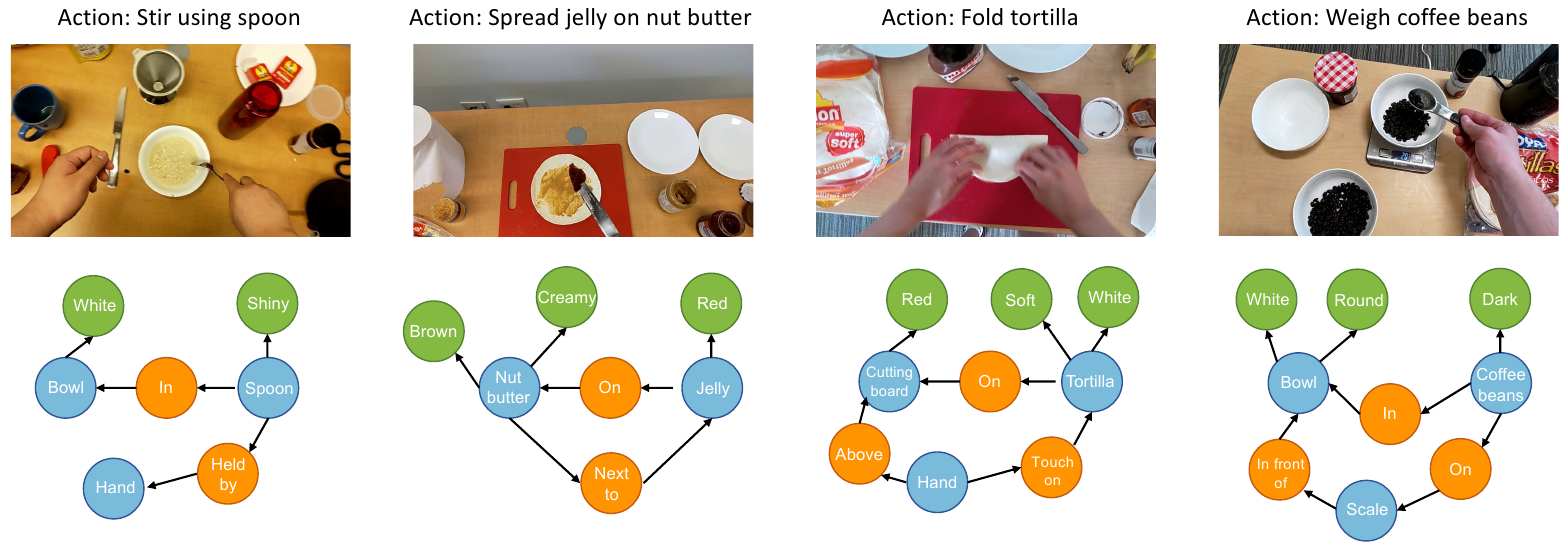}
    \caption{Examples of action scene graphs built upon effect frames.}
    \label{fig:vis_graph}
\end{figure}

Figure \ref{fig:vis_graph} demonstrates examples of action scene graph build upon GPT-4o recognition and parsed by Algorithm \ref{alg:scene_graph_construct}. They captures attributes and spatial relationship of objects in a structured manner.

\subsection{Action Segment Frame Scoring}
\begin{figure}[t]
    \centering
    \includegraphics[width=\linewidth]{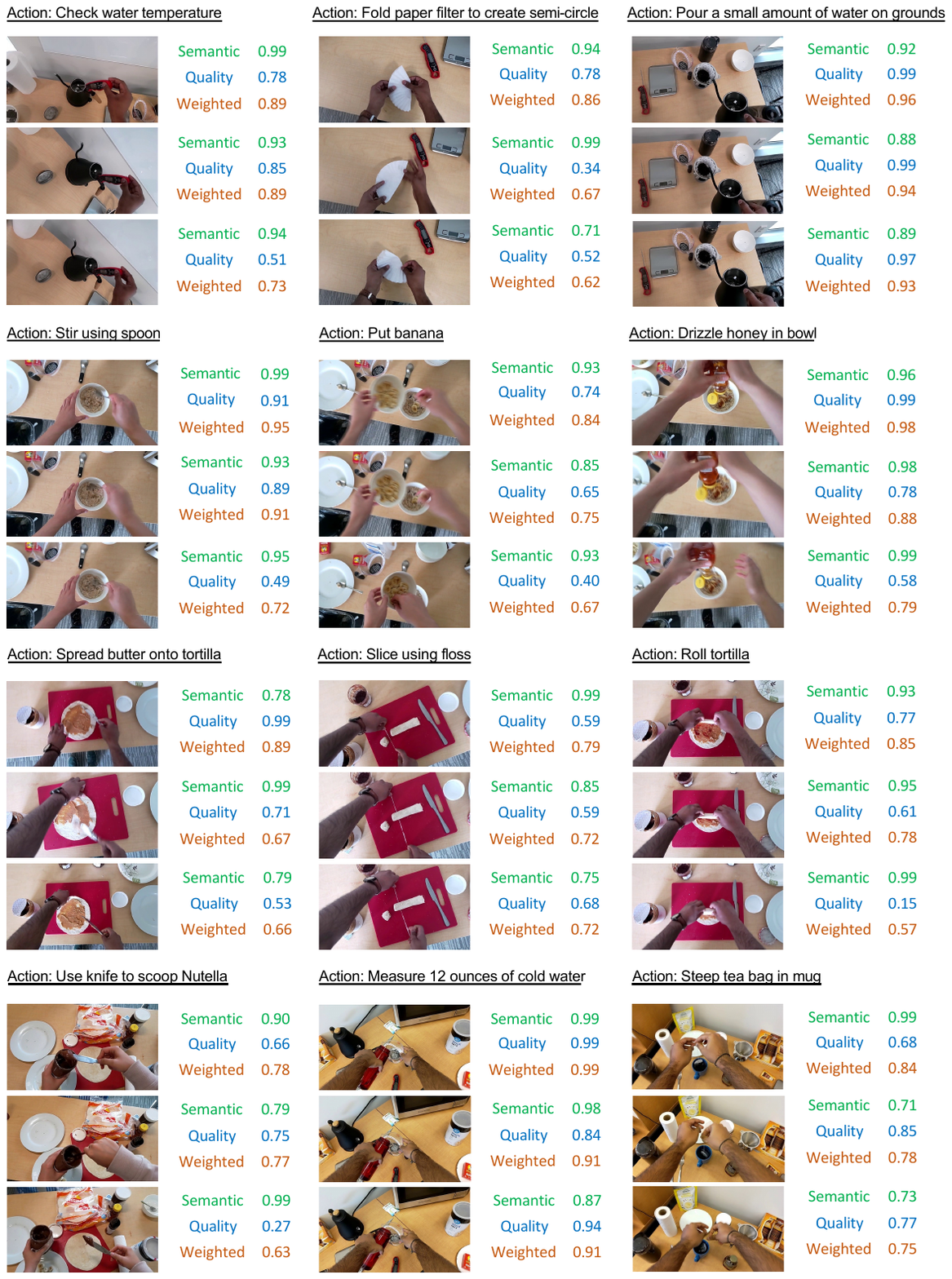}
    \caption{Examples of scores by effect frame sampling for different actions.}
    \label{fig:vis_frame_sample}
\end{figure}

Figure~\ref{fig:vis_frame_sample} shows examples of the top-3 video frames with the highest weighted scores in effect-frame sampling. In our implementation, semantic and quality scores are averaged to produce the final weights. Visual inspection indicates that these frames, which are often consecutive in the video, contain highly similar content. Empirically, we find that using the top-$K$ frames instead of only the top-1 provides negligible performance gains while substantially increasing annotation and training costs. When $K$ is larger (e.g., $K=5$), lower-scoring frames may introduce noisy or mismatched content, leading to reduced mistake-detection performance. Therefore, we select only the highest-scoring frame as the effect frame in our sampling strategy.

\clearpage
\section{Further Discussion}
\label{section:a4}

\paragraph{Dependence on object detection and scene graphs.} The performance of our framework remains tied to the accuracy of object detection and scene graph generation. In egocentric and cluttered environments, objects are frequently occluded, blurred, or partially visible, leading to unstable detection and relationship modeling, which in turn degrades the reliability of error detection. Future directions include incorporating stronger spatiotemporal representations, such as 3D or 4D modeling, together with multimodal cues, such as depth maps or point clouds, to improve robustness under occlusion and enhance object localization and relational reasoning.

\paragraph{Spatial-temporal modeling.} Current effect modeling focuses on single-step actions and does not explicitly capture long-range dependencies across multiple steps. In real procedural tasks, small early deviations can accumulate and lead to significant downstream errors. Future directions include modeling cross-step dependencies through hierarchical task graphs, causal reasoning structures, or program-level state transition models, thereby extending the framework to long-horizon, multi-step reasoning and enabling more comprehensive detection of complex procedural errors.

\paragraph{Supervision of effect modeling.} Our framework currently relies on multimodal supervision generated by large pretrained models, which may hinder generalization to new domains or resource-constrained settings. To reduce this dependency, future work may explore weakly supervised or self-supervised signals, for example, leveraging temporal consistency, action outcome comparisons, or environment state transitions to automatically construct supervision, thereby improving scalability in broader domains and scenarios.

\paragraph{Dependence on large models.} Our method relies on large models to generate pseudo-labels or auxiliary signals for action effect modeling, yet their outputs can be stochastic and inconsistent. For instance, in experiments, we observed that an identical effect frame may yield slightly different spatial relationship descriptions across runs, and such noise can propagate into learned effect representations and impact training stability. Potential mitigation strategies include adopting open-source models with reproducible checkpoints and fixed random seeds, as well as introducing consistency constraints, cross-model voting, or automatic filtering mechanisms to reduce noise propagation.

\paragraph{Interpretability of mistakes.} Most existing methods, including ours, formulate error detection as binary classification, which provides limited insight into the underlying causes of errors. Enhancing interpretability therefore represents an important research direction. One possible solution is to integrate LLMs or LVLMs to generate human-understandable explanations, improving the transparency and usability of error detection systems.

\paragraph{Potential of world models.} World models have recently emerged as an active research direction, as they capture environment dynamics and predict future scenes based on historical observations. These capabilities make them promising for procedural video mistake detection, particularly for online and one-class classification (OCC) settings. Future directions include predicting future states in pixel or latent space conditioned on past observations and comparing them with actual outcomes to identify inconsistencies, which correspond to procedural errors.
\section{Use of Large Language Model}
\label{section:a5}
In this work, large language models (LLMs) were employed solely for English translation and language polishing, with the goal of improving fluency and clarity of the manuscript so that the contributions can be presented to readers more effectively. LLMs were \textbf{not} used for idea generation, conceptualization, or any other aspect of the research.
\end{document}